\documentclass[11pt]{article}

\usepackage[preprint]{acl}

\usepackage{times}
\usepackage{latexsym}
\usepackage{amsmath}
\usepackage[T1]{fontenc}
\usepackage[utf8]{inputenc}
\usepackage{microtype}
\usepackage{inconsolata}
\usepackage{graphicx}

\usepackage{comment}
\usepackage{tikz}
\usetikzlibrary{positioning,calc,arrows.meta,fit,backgrounds}
\usepackage{booktabs}
\usepackage{array}
\usepackage[table,dvipsnames]{xcolor}
\usepackage{enumitem}
\usepackage{arydshln}

\title{No \textsc{pun} Intended:\\
Plausible Unknown Names for Person-Centred LLM Evaluation}

\author{
Dimitri Staufer\textsuperscript{1} \quad
David Hartmann\textsuperscript{1,2} \quad
Ibrahim Baroud\textsuperscript{3,4} \\
\textsuperscript{1}Technische Universität Berlin, Berlin, Germany \\
\textsuperscript{2}Weizenbaum Institute for the Networked Society, Berlin, Germany \\
\textsuperscript{3}Quality \& Usability Lab, Technische Universität Berlin, Berlin, Germany \\
\textsuperscript{4}German Research Center for Artificial Intelligence (DFKI), Berlin, Germany
}

\begin{document}
\maketitle

\begin{abstract}
Person names are widely used as prompt variables in LLM evaluations of factuality, privacy leakage, bias and abstention, but when a name’s evidential status is uncontrolled, measurements may conflate memorisation, retrieval, name priors and wrong-person attribution. We operationalise an unknown name as one with plausible First-Last form, no indexed full-name evidence, and no ambiguity signals under a documented validation run, and introduce \textsc{pun} (Plausible Unknown Names), a protocol for constructing and validating such names, combining Wikidata-derived components, web-enabled LLM screening, and controlled search revalidation. We report acceptance rate, reproducibility, ablations, and a 204-participant human study, finding accepted names are more name-like than controls while participants recover person evidence in only 3\% of cases. We release 300 names with comparison controls.
\end{abstract}

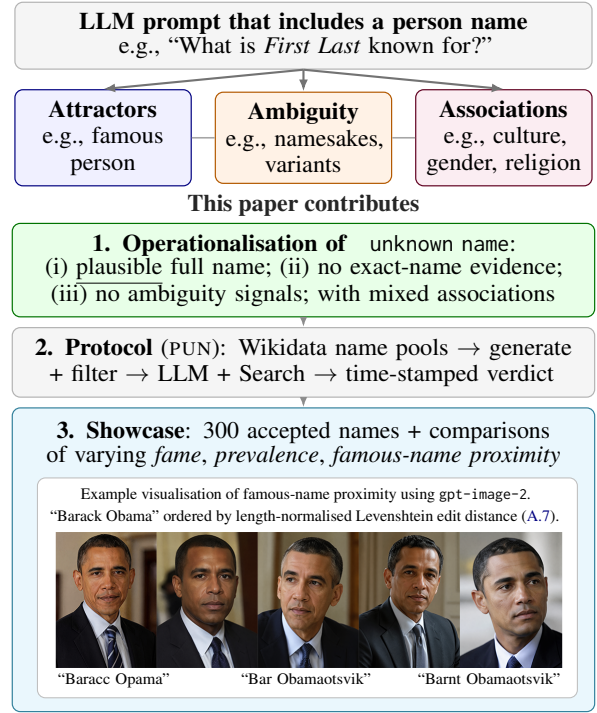
\begin{figure}[t]
\centering
\small

\def\stripheight{1.35cm}
\def\rowsep{2pt}

\begin{tikzpicture}[
    font=\small,
    base/.style={
        draw,
        rounded corners=3pt,
        align=center,
        inner sep=4pt,
        line width=.45pt
    },
    topbox/.style={
        base,
        fill=gray!8,
        draw=gray!60,
        text width=7.35cm,
        minimum height=.72cm
    },
    risk/.style={
        base,
        text width=2.05cm,
        minimum height=.86cm
    },
    contrib/.style={
        base,
        text width=7.35cm,
        minimum height=.72cm,
        inner xsep=5pt,
        inner ysep=4pt
    },
    arr/.style={
        -{Latex[length=1.6mm]},
        thick,
        draw=black!55
    },
    seqarr/.style={
        -{Latex[length=1.35mm]},
        semithick,
        draw=black!45
    },
    signalconn/.style={
        semithick,
        draw=black!35
    },
    usecase/.style={
    base,
    font=\scriptsize,
    text width=2.18cm,
    minimum height=.72cm,
    inner xsep=3pt,
    inner ysep=3pt,
    fill=yellow!7,
    draw=orange!55!black
    }
]

\node[topbox] (prompt) at (0,0) {
\textbf{LLM prompt that includes a person name}\\[-1pt]
e.g., ``What is \emph{First Last} known for?''
};

\node[risk, fill=blue!6, draw=blue!50!black] (attractors) at (-2.65,-1.35) {
\textbf{Attractors}\\
e.g., famous person
};

\node[risk, fill=orange!10, draw=orange!70!black] (ambiguity) at (0,-1.35) {
\textbf{Ambiguity}\\
e.g., namesakes, variants
};

\node[risk, fill=purple!7, draw=purple!50!black] (assoc) at (2.65,-1.35) {
\textbf{Associations}\\
e.g., culture, gender, religion
};

\draw[signalconn] (attractors.east) -- (ambiguity.west);
\draw[signalconn] (ambiguity.east) -- (assoc.west);

\draw[arr] (prompt.south) -- (attractors.north);
\draw[arr] (prompt.south) -- (ambiguity.north);
\draw[arr] (prompt.south) -- (assoc.north);

\node[
    font=\small\bfseries,
    fill=white,
    inner sep=1.5pt,
    text=black!85
] (contriblabel) at (0,-2.24) {
This paper contributes
};

\node[
    contrib,
    fill=green!9,
    draw=green!45!black
] (defn) at (0,-3.1) {
\textbf{1. Operationalisation of}\quad
\texttt{unknown name}:\\[-1pt]
(i) \underline{plausible} full name; (ii) no exact-name evidence; (iii) no ambiguity signals; with mixed associations
};

\node[
    contrib,
    fill=gray!8,
    draw=gray!60
] (protocol) at (0,-4.32) {
\textbf{2. Protocol} (\textsc{pun}): 
Wikidata name pools
$\rightarrow$ generate + filter
$\rightarrow$ LLM + Search
$\rightarrow$ time-stamped verdict
};

\node[
    contrib,
    fill=cyan!7,
    draw=cyan!55!black,
    minimum height=0cm,
    inner ysep=5pt
] (dataset) at (0,-6.93) {
\begin{minipage}{7.05cm}
\centering

\makebox[\linewidth][c]{%
\textbf{3. Showcase}: 
300 accepted names + comparisons%
}\\[-1pt]
{\small of varying \emph{fame}, \emph{prevalence}, \emph{famous-name proximity}}\\[3pt]

\begin{tikzpicture}
    \node[
        draw=black!25,
        rounded corners=3pt,
        line width=0.4pt,
        fill=white,
        inner xsep=4pt,
        inner ysep=2pt
    ] (frame) {
        \begin{minipage}{6.78cm}
        \centering

        \vspace*{2pt}%
        {\tiny
        Example visualisation of famous-name proximity using \texttt{gpt-image-2}.\\[-2pt]
        ``Barack Obama'' ordered by length-normalised Levenshtein edit distance
        (\ref{sec:appendix-comparison-names}).%
        }\\[2pt]

        \hspace*{2pt}%
        \begin{tikzpicture}
            \node[inner sep=0pt] (img) {
                \includegraphics[
                    width=\dimexpr\linewidth-2pt\relax,
                    height=1.82cm,
                    keepaspectratio
                ]{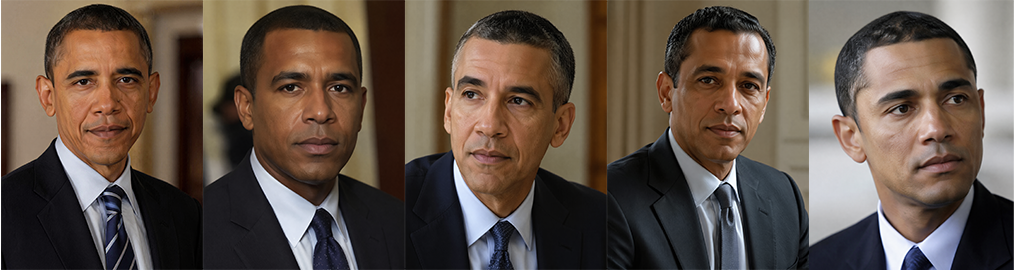}
            };

            \coordinate (face1) at ($(img.south west)!0.10!(img.south east)$);
            \coordinate (face3) at ($(img.south west)!0.50!(img.south east)$);
            \coordinate (face5) at ($(img.south west)!0.90!(img.south east)$);

            \node[
                font=\tiny,
                align=center,
                text width=1.45cm,
                anchor=north,
                inner sep=2pt,
                outer sep=0pt
            ] (lab1) at ($(face1)+(0.1cm,0)$) {%
                ``Baracc Opama''%
            };

            \node[
                font=\tiny,
                align=center,
                text width=1.75cm,
                anchor=north,
                inner sep=2pt,
                outer sep=0pt
            ] (lab3) at ($(face3)+(-0.08cm,0)$) {%
                ``Bar Obamaotsvik''%
            };

            \node[
                font=\tiny,
                align=center,
                text width=3.05cm,
                anchor=north,
                inner sep=2pt,
                outer sep=0pt
            ] (lab5) at ($(face5)+(-0.38cm,0)$) {%
                ``Barnt Obamaotsvik''%
            };

        \end{tikzpicture}
        \end{minipage}
    };
\end{tikzpicture}

\end{minipage}
};

\draw[seqarr] (defn.south) -- (protocol.north);
\draw[seqarr] (protocol.south) -- (dataset.north);

\end{tikzpicture}

\caption{
Paper motivation and contribution overview.
}
\label{fig:visual-operational-unknownness}
\end{figure}


\section{Introduction}

Person names appear across many large language model (LLM) evaluation settings. They are used as targets for biography generation \citep{min-etal-2023-factscore}, prompts for personal-data auditing \citep{staufer-etal-2026-human,tan-etal-2025-personabench}, cues in hiring-decision audits \citep{an-etal-2024-large}, possible memorised identifiers in privacy leakage tests \citep{carlini2021extracting,huang-etal-2022-large,keum2025private}, spans to be replaced in anonymisation and pseudonymisation \citep{lison-etal-2021-anonymisation,eder-etal-2019-de,yermilov-etal-2023-privacy}, inputs for evaluating abstention when evidence is insufficient \citep{madhusudhan-etal-2025-llms,wen-etal-2025-know}, and elements of LLM-generated synthetic examples \citep{wang-etal-2023-self-instruct,li-etal-2023-synthetic,long-etal-2024-llms}. Across these settings, a name is often treated as a convenient prompt variable: something that can be inserted, replaced, perturbed, or compared across conditions.

Yet person names are not neutral placeholders. A full-name string can simultaneously act as a key to public evidence about a referent, as in ``Barack Obama''; an ambiguous identifier shared by multiple people, as in ``John Williams''; a form linked to nearby spellings or reordered variants, as in ``Stephen King'' versus ``Steven King''; and a signal of culture, gender, ethnicity, religion, or class. These roles have been studied separately in work on web-based person-name disambiguation \citep{artiles-etal-2007-semeval,balog2009resolving,delgado2018survey} and name-conditioned social or demographic inference \citep{jeoung-etal-2023-examining,gautam-etal-2024-stop,pawar-etal-2025-presumed}. As Figure~\ref{fig:visual-operational-unknownness} illustrates, several signals can attach to the same prompted full-name string, making measurements in person-centred LLM evaluations difficult to interpret when the string’s evidential status is uncontrolled.

For example, when an LLM answers ``What is \emph{First Last} known for?'' in a biography evaluation, recommends whether to interview \emph{First Last} in a hiring audit, produces personal details for \emph{First Last} in a privacy test, or refuses to answer a question such as ``Has \emph{First Last} been arrested?'', the behaviour may have several causes. It may reflect indexed public evidence, memorised training data, retrieval at inference time, name-conditioned social priors, wrong-person conflation, or unsupported generation. Conversely, when a model abstains, it may be appropriately uncertain, may have failed to retrieve available evidence, or may be following a generic safety pattern unrelated to the name itself. Using names with different amounts and kinds of public evidence can therefore help tease apart these possible explanations. Without such control, measurements of factuality, privacy leakage, bias, and abstention can conflate distinct sources of model behaviour.

\paragraph{Approach.} We address this problem by developing a flexible protocol for generating operationally unknown names. We define \emph{operational unknownness} for person-name prompts using a protocol we call \textsc{pun} (Plausible Unknown Names). It constructs, screens, and revalidates plausible person-name prompts. Here, ``plausible'' means a two-component \emph{First Last} string built from attested given-name and surname components and passing explicit form constraints. It does not mean culturally universal, naturally acceptable, or legally valid as a name. The protocol relies on Wikidata for candidate generation and uses a web-enabled LLM and search engine for verification. The term ``operational'' is deliberate. We do not claim non-existence of a person or that no private record contains the name. Nor do we claim that future search results will remain empty. We claim only that the name satisfies a conservative, time-bounded evidential condition that can be inspected, challenged, and revalidated.

\paragraph{Contributions.}

This paper makes three contributions: (1) a practical operationalisation of name unknownness, specifying both
the acceptance criterion and diagnostics for borderline cases; (2) \textsc{pun}, an auditable protocol for generating and validating plausible unknown names; and (3) a showcase resource of 300 accepted names with comparison controls, validation traces, reproducibility diagnostics, protocol ablations, and a 204-participant human agreement audit.

\section{Operational Unknownness}
\label{sec:definition}

We use \emph{operationally unknown} for the binary evidential status of a plausible full-name string under a specified validation run. Let \(n\) be the full-name string used in an evaluation prompt, and let \(PUN_t\) denote the validation protocol run at time \(t\). The protocol \(PUN_t\) specifies the retrieval systems, settings, query forms, and decision rules used to search for indexed public evidence. Section~\ref{sec:protocol} describes the concrete instantiation used in this paper.

A candidate is accepted as operationally unknown under \(PUN_t\) if (i) it is a plausible two-component \emph{First Last} string, (ii) \(PUN_t\) finds no indexed public evidence that the exact full name, or a configured equivalent, refers to a person, and (iii) \(PUN_t\) finds no full-name-level ambiguity signal, such as a nearby spelling, reordered form, search correction, famous-name attractor, or unclear non-person referent. For example, \emph{Leora Vandeskel} would be rejected if the run found a trace for that exact name, a reordered form such as \emph{Vandeskel Leora}, a similar spelling such as \emph{Liora Vandeskel}, or a non-person entity such as ``Vandeskel Corp''.

\section{Background and Related Work}
\label{sec:background}
\label{sec:related-work}

\subsection{Name Form, Resources, and Plausibility}
\label{sec:background-name-form}
\label{sec:rw-name-resources}

\textsc{pun} uses \emph{plausible name} in a narrow operational sense: a candidate has \emph{First Last} form, is built from attested given-name and surname components, and passes explicit form checks. This is less ambitious than sociolinguistic, legal, demographic, or community-specific plausibility. Personal names vary in morphology, orthography, frequency, conventionality, cultural distribution, and social meaning \citep{vanlangendonck2007theory,motschenbacher2020corpus}. Naming systems also vary in order, number of components, inherited surnames, patronymics, mononyms, particles, multiple surnames, and script conventions \citep{lawson2016personal,ndlovu2023personal}. Existing name resources and pseudonymisation work usually start from observed spans: ParaNames derives Wikidata entity names for multilingual NER, linking, translation, and transliteration \citep{saleva-lignos-2024-paranames}, while anonymisation and pseudonymisation systems replace sensitive person-name spans with masks or surrogates \citep{lison-etal-2021-anonymisation,eder-etal-2019-de,yermilov-etal-2023-privacy,szawerna-etal-2024-pseudonymization,szawerna-etal-2026-fill}. \textsc{pun} instead constructs full-name prompts that retain name-like form while making their indexed full-name evidence auditable.

\subsection{Full Names as Ambiguous Evidence Keys}
\label{sec:background-name-ambiguity}
\label{sec:rw-disambiguation-evidence}

A full name is not a reliable identifier. The same string can refer to multiple people, and one person can appear under multiple spellings, scripts, transliterations, aliases, abbreviations, reordered forms, or other variants. Web-based person-name disambiguation is built around this problem: systems must separate namesakes while reconciling variant forms \citep{artiles-etal-2007-semeval,balog2009resolving,delgado2018survey,zhagorina2018personal}. Absence of search evidence is also not absolute, since search-engine coverage and retrievability can change over time, and hit-count estimates can be approximate, unstable, and provider-dependent \citep{barilan1999search,uyar2009investigation,sanchez2018survey}.

\subsection{Names in LLM Evaluation}
\label{sec:background-names-llm-eval}
\label{sec:rw-names-llm-evaluation}

LLM evaluations use names in biography and factuality evaluation \citep{min-etal-2023-factscore}, privacy and memorisation audits \citep{carlini2021extracting,huang-etal-2022-large,keum2025private,staufer-etal-2026-human,tan-etal-2025-personabench}, anonymisation and pseudonymisation settings \citep{lison-etal-2021-anonymisation,eder-etal-2019-de,yermilov-etal-2023-privacy}, hiring and employment audits \citep{an-etal-2024-large,wang-etal-2024-jobfair}, bias and social-inference studies \citep{dearteaga2019bias,romanov2019name,jeoung-etal-2023-examining,an-etal-2025-mutual,pawar-etal-2025-presumed}, and abstention or unanswerability tests \citep{rajpurkar-etal-2018-know,madhusudhan-etal-2025-llms,wen-etal-2025-know}. Nonword or malformed-input probes avoid public-person evidence by moving away from plausible names \citep{ramakrishna-etal-2023-invite,vitevitch2025examining}. \textsc{pun} occupies the middle ground: the input remains a plausible full-name prompt, but indexed full-name evidence and full-name-level ambiguity signals are controlled by construction. When a model produces person-specific facts for such a name, those facts are unsupported relative to the documented validation run of \textsc{pun} rather than attributable to indexed full-name evidence found by the protocol.

\begin{figure*}[t]
\centering
\small
\resizebox{0.99\textwidth}{!}{%
\begin{tikzpicture}[
    font=\small,
    node distance=0.44cm,
    stage/.style={
        draw,
        rounded corners=3pt,
        align=center,
        text width=2.35cm,
        minimum height=1.05cm,
        inner sep=4pt,
        fill=gray!7,
        draw=gray!60
    },
    stagewiki/.style={
        draw,
        rounded corners=3pt,
        align=center,
        text width=2.65cm,
        minimum height=1.05cm,
        inner sep=4pt,
        fill=gray!7,
        draw=gray!60
    },
    example/.style={
        draw,
        rounded corners=3pt,
        align=center,
        text width=2.52cm,
        minimum height=.62cm,
        inner sep=3pt,
        fill=yellow!8,
        draw=orange!55!black,
        font=\scriptsize
    },
    pass/.style={
        draw,
        rounded corners=3pt,
        align=center,
        text width=2.35cm,
        minimum height=.62cm,
        inner sep=3pt,
        fill=green!7,
        draw=green!45!black,
        font=\scriptsize
    },
    reject/.style={
        draw,
        rounded corners=3pt,
        align=center,
        text width=2.35cm,
        minimum height=.68cm,
        inner sep=3pt,
        fill=red!5,
        draw=red!55!black,
        font=\scriptsize
    },
    accept/.style={
        draw,
        rounded corners=3pt,
        align=center,
        text width=1.75cm,
        minimum height=.56cm,
        inner sep=3pt,
        fill=green!8,
        draw=green!45!black,
        font=\scriptsize
    },
    phasebox/.style={
        draw=black!16,
        rounded corners=5pt,
        fill=black!1,
        inner xsep=5pt,
        inner ysep=4pt
    },
    phaselabel/.style={
        font=\scriptsize\bfseries,
        text=black!45,
        fill=white,
        inner xsep=2pt,
        inner ysep=4pt
    },
    arr/.style={
        -{Latex[length=1.7mm]},
        thick,
        draw=black!55
    }
]

\node[stagewiki] (pools) {
    \textbf{Wikidata}\\
    \textbf{name pools}\\
    \emph{birthplace} (\emph{country})
};

\node[stage, right=.22cm of pools] (gen) {
    \textbf{Generation}\\
    sample one \emph{given}\\
    and one \emph{surname}
};

\node[stage, right=.50cm of gen] (local) {
    \textbf{Format}\\
    \textbf{Validation}\\
    \emph{First Last}; no titles; no suffixes; Latin script
};

\node[stage, right=.50cm of local] (web) {
    \textbf{Web-enabled}\\
    \textbf{LLM queries}\\
    \texttt{Who is \{name\}?}
};

\node[stage, right=.22cm of web] (search) {
    \textbf{Controlled}\\
    \textbf{Web Search}\\
    exact name; variants
};

\node[accept, right=.32cm of search] (accepted) {
    \textbf{Operationally}\\
    \textbf{unknown}
};

\begin{scope}[on background layer]
\node[
    phasebox,
    fit=(pools)(gen),
    label={[phaselabel]above:{1. Candidate generation}}
] {};

\node[
    phasebox,
    fit=(local),
    label={[phaselabel]above:{2. Local checks}}
] {};

\node[
    phasebox,
    fit=(web)(search),
    label={[phaselabel]above:{3. Web-facing validation}}
] {};
\end{scope}

\draw[arr] (pools) -- (gen);
\draw[arr] (gen) -- (local);
\draw[arr] (local) -- (web);
\draw[arr] (web) -- (search);
\draw[arr] (search) -- (accepted);

\node[example, text width=5.35cm, minimum height=1.18cm] (expools)
at ($(pools.south)!0.5!(gen.south)+(0,-1.1cm)$) {
    \textbf{Examples}\\
    \emph{Givens}: Adeline (France); Afr\^anio (Brazil); Anjali (India); Akifumi (Japan)\\
    \emph{Surnames}: S\"ollig (Germany); Abad\'ia (Spain); Adekunle (Nigeria); Antonenko (Ukraine)
};

\node[example, text width=5.35cm, below=.23cm of expools] (exgen) {
    \emph{given} (Brazil) + \emph{surname} (Ukraine) \(\rightarrow\)\\
    ``Afr\^anio Antonenko''
};

\draw[arr] (expools.south) -- (exgen.north);

\node[pass, below=.35cm of local] (passlocal) {
    ``Didier-Jules S\"ollig''
};

\node[reject, below=.35cm of passlocal] (faillocal) {
    ``Dr. Zoe van Delk''\\
    $\rightarrow$ has title
};

\draw[arr] (local.south) -- (passlocal.north);
\draw[arr] (passlocal.south) -- (faillocal.north);

\node[pass, below=.45cm of web] (passweb) {
    ``Moffat Bourde''
};

\node[reject, below=.35cm of passweb] (failweb) {
    ``Mehmet Novick''\\
    $\rightarrow$ found, similar, or attractor
};

\draw[arr] (web.south) -- (passweb.north);
\draw[arr] (passweb.south) -- (failweb.north);

\node[pass, below=.35cm of search] (passsearch) {
    ``Pinsach Nibbrig''
};

\node[reject, below=.35cm of passsearch] (failsearch) {
    ``Ironside Birlik''\\
    $\rightarrow$ full-name hit, near trace, or search hint
};

\draw[arr] (search.south) -- (passsearch.north);
\draw[arr] (passsearch.south) -- (failsearch.north);

\end{tikzpicture}%
}
\caption{
Overview of \textsc{pun}'s three stages: candidate generation, local format checks, and web-facing validation.
}
\label{fig:protocol-pipeline}
\end{figure*}
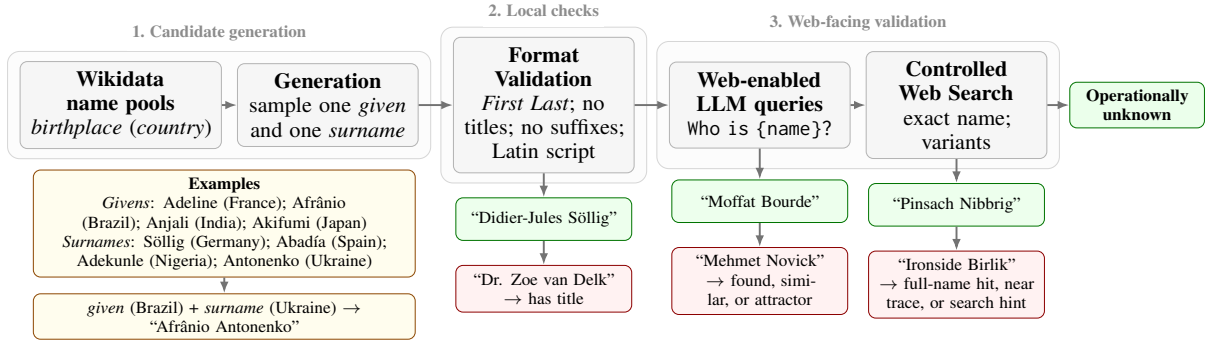

\section{The \textsc{pun} Protocol}
\label{sec:protocol}

Given candidate two-component name strings, \textsc{pun} checks whether they satisfy the operational-unknownness criterion from Section~\ref{sec:definition}: no indexed public full-name evidence and no full-name-level ambiguity signals under a documented validation run. The released showcase in Section~\ref{sec:pun-resource} is one run of this protocol, \(PUN_t\), with fixed sources, providers, prompts, query templates, thresholds, and timestamps. Appendix~\ref{sec:appendix-protocol-details} gives the implementation details needed for replication.

As Figure~\ref{fig:protocol-pipeline} shows, the protocol has three stages: (1) candidate generation, (2) local format checks, and (3) web-facing validation. Candidate generation builds plausible \emph{First Last} strings from attested Wikidata name components. Local checks remove malformed or out-of-scope strings before web-facing validation. The web-facing stages then assign the evidential verdict, with a web-enabled LLM for broad triage and controlled web searches as the auditable basis for acceptance or rejection.

The two web-facing stages are deliberately separated. The web-enabled LLM can surface exact referents, spelling or order variants, fictional characters, organisations, famous-name attractors, and other nearby matches that fixed search templates may miss. It is not the evidential basis for acceptance, because its retrieval path, query expansion, and ranking decisions are not fully observable.

\subsection{Name Sources and Candidate Generation}
\label{sec:protocol-generation}

We use full names from Wikidata human entities as a source of attested name surface forms, rather than generating pronounceable but possibly fictional or parodic strings. We parse entity labels into given-name and surname positions and derive a country key by mapping the person's recorded birthplace to a country\footnote{The source-country key is used only for sampling and audit. It is not assigned to generated names and is not interpreted as nationality, ethnicity, language, or demographic status.}. Candidates are sampled by combining one given-name component and one surname component into a \emph{First Last} string (Figure~\ref{fig:protocol-pipeline}.1).

Sampled candidates are filtered by form, script, length, frequency, and duplicate checks. Candidates must have exactly two components, no titles or suffixes, Latin-script form with diacritics retained, and a bounded total length. We also exclude overly common components and require each retained component to appear in multiple source records and to have native-language metadata. These filters reduce malformed strings, parsing artefacts, and combinations that are unlikely to survive evidence screening (Figure~\ref{fig:protocol-pipeline}.2). They make plausibility a documented form condition rather than a claim that a name is culturally representative, legally valid, or natural in every naming system.

\subsection{Web-facing Validation}
\label{sec:protocol-validation}

After generation and local checks, candidates are first screened with a web-enabled LLM using a person-identification query (``Who is \{name\}?''). The response is adjudicated into one of four labels:

\begin{itemize}[noitemsep, topsep=0pt]
    \item \texttt{found}: the exact full name is tied to concrete biographical detail.
    \item \texttt{ambiguous}: the response points to a similar spelling, reordered form, partial-token person, fictional character, organisation, place, work, product, or other nearby referent.
    \item \texttt{fail}: the response is policy-like, unusable, inconclusive, or otherwise failed.
    \item \texttt{no\_info}: the response reports no information about the exact full-name string and mentions no alternate person or entity.
\end{itemize}

Only candidates labelled \texttt{no\_info} proceed to controlled web search, where we check the exact full name and configured variants, including quoted and unquoted forms, reversed order, ASCII-folded forms for names with diacritics, and bounded spelling or phonetic variants. These variants are intended to catch cases where a bare exact query misses evidence but nearby forms, search corrections, or reordered strings do.

A web search candidate is rejected as \texttt{found} if the exact or configured full-name form appears in a person-like context in a search-result title, snippet, URL, knowledge-panel field, or page content. We operationalise person-like context with fixed biographical terms, person-oriented domains, and knowledge-panel type cues. If the name appears only in a non-person or unclear context, or if search corrections and configured variants point to a nearby full-name form, the candidate is rejected as \texttt{ambiguous}. A name is accepted as \textbf{operationally unknown} only when controlled search finds neither indexed public full-name evidence nor full-name-level ambiguity signals.


\subsection{Comparison-Name Construction}
\label{sec:protocol-comparison}

Alongside the operationally unknown names, we construct comparison names for boundary cases around operational unknownness. They let us (i) compare accepted unknown names with names of public people whose \emph{fame} and exact-name widespreadness varies, and (ii) use strings that are close to public-person names at controlled surface distances (e.g., \emph{Mike Myers} \(\rightarrow\) \emph{Mi\underline{c}e Myers} \(\rightarrow\) \emph{Mike\underline{uela} \underline{Olou}}) to measure how strongly models map unfamiliar name-like strings onto existing names as surface similarity decreases.

Public-person controls are sampled from Wikidata human entities. We use page views, article length, language count, and sitelink count as a reproducible famousness proxy, and combine this with quoted exact-name search-result counts as a proxy for full-name widespreadness. This separates a person's public prominence from the web footprint of the literal full-name string.

Famous-name distance controls are build by perturbing Wikidata-linked public-person names in 12 increasingly distant layers, from close orthographic edits to suffix attachments and fragmentary recombinations. Appendix~\ref{sec:appendix-comparison-names} reports the release strata and construction details.

\section{The \textsc{pun} Resource}
\label{sec:pun-resource}

The released resource is a showcase set from one documented run of \textsc{pun}, not a dataset of permanently unknown people. It contains 300 full names accepted as operationally unknown under \textsc{pun}, together with 300 public-person comparison names of varying fame and full-name prevalence, and 12 x 300 = 3,600 famous-name distance controls. The entire source code and showcase resource are publicly available under CC BY-NC 4.0.\footnote{\url{https://anonymous.4open.science/r/PUN}}

\subsection{Name Candidate Generation}
\label{sec:resource-candidate-generation}

We use Wikidata human labels as an auditable source of attested name components, not as a representative sample of global naming practices. For this run, we processed the Wikidata dump from May 2026. It yielded 13.5M human entities, of which 4.17M entered the source pool after name parsing and birthplace-country mapping. Tokenising this pool produced 356,558 distinct given-name components and 989,563 surname components. After local filters, 8,404 given names and 30,576 surnames remained eligible (details in Appendix~\ref{sec:appendix-candidate-generation}).

We sample names components by source-country rather than by raw pool size to reduce geographical skew, and disallow same-country recombinations to lower the chance of reconstructing attested within-country full names. The generated evaluation-scope candidates span 545 observed source-country pairs. Appendix Figures~\ref{fig:wikidata-filtered-recombination-country-pairs-full} and~\ref{fig:wikidata-candidate-country-pairs-full} show the full filtered-slot and generated-candidate country-pair distributions.

\begin{figure}[t]
\centering
\begin{tikzpicture}[x=1cm,y=1cm,
  lab/.style={font=\scriptsize, text=black!75},
  axislab/.style={font=\scriptsize, text=black!70},
  num/.style={font=\tiny, text=black!85},
  numdark/.style={font=\tiny, text=white},
]
\node[axislab, anchor=east] at (-.08,0.310) {Other};
\path[fill=blue!55] (0.000,0.000) rectangle (0.620,0.620);
\draw[white, line width=.15pt] (0.000,0.000) rectangle (0.620,0.620);
\node[numdark] at (0.310,0.310) {17};
\path[fill=blue!49] (0.620,0.000) rectangle (1.240,0.620);
\draw[white, line width=.15pt] (0.620,0.000) rectangle (1.240,0.620);
\node[numdark] at (0.930,0.310) {12};
\path[fill=blue!51] (1.240,0.000) rectangle (1.860,0.620);
\draw[white, line width=.15pt] (1.240,0.000) rectangle (1.860,0.620);
\node[numdark] at (1.550,0.310) {13};
\path[fill=blue!47] (1.860,0.000) rectangle (2.480,0.620);
\draw[white, line width=.15pt] (1.860,0.000) rectangle (2.480,0.620);
\node[numdark] at (2.170,0.310) {10};
\path[fill=blue!45] (2.480,0.000) rectangle (3.100,0.620);
\draw[white, line width=.15pt] (2.480,0.000) rectangle (3.100,0.620);
\node[numdark] at (2.790,0.310) {9};
\path[fill=blue!39] (3.100,0.000) rectangle (3.720,0.620);
\draw[white, line width=.15pt] (3.100,0.000) rectangle (3.720,0.620);
\node[numdark] at (3.410,0.310) {6};
\path[fill=blue!78] (3.720,0.000) rectangle (4.340,0.620);
\draw[white, line width=.15pt] (3.720,0.000) rectangle (4.340,0.620);
\node[numdark] at (4.030,0.310) {69};
\node[axislab, anchor=east] at (-.08,0.930) {Brazil};
\path[fill=blue!29] (0.000,0.620) rectangle (0.620,1.240);
\draw[white, line width=.15pt] (0.000,0.620) rectangle (0.620,1.240);
\node[num] at (0.310,0.930) {3};
\path[fill=blue!18] (0.620,0.620) rectangle (1.240,1.240);
\draw[white, line width=.15pt] (0.620,0.620) rectangle (1.240,1.240);
\node[num] at (0.930,0.930) {1};
\path[fill=blue!25] (1.240,0.620) rectangle (1.860,1.240);
\draw[white, line width=.15pt] (1.240,0.620) rectangle (1.860,1.240);
\node[num] at (1.550,0.930) {2};
\path[fill=blue!6] (1.860,0.620) rectangle (2.480,1.240);
\draw[white, line width=.15pt] (1.860,0.620) rectangle (2.480,1.240);
\node[num] at (2.170,0.930) {0};
\path[fill=blue!18] (2.480,0.620) rectangle (3.100,1.240);
\draw[white, line width=.15pt] (2.480,0.620) rectangle (3.100,1.240);
\node[num] at (2.790,0.930) {1};
\path[fill=blue!18] (3.100,0.620) rectangle (3.720,1.240);
\draw[white, line width=.15pt] (3.100,0.620) rectangle (3.720,1.240);
\node[num] at (3.410,0.930) {1};
\path[fill=blue!45] (3.720,0.620) rectangle (4.340,1.240);
\draw[white, line width=.15pt] (3.720,0.620) rectangle (4.340,1.240);
\node[numdark] at (4.030,0.930) {9};
\node[axislab, anchor=east] at (-.08,1.550) {Germany};
\path[fill=blue!6] (0.000,1.240) rectangle (0.620,1.860);
\draw[white, line width=.15pt] (0.000,1.240) rectangle (0.620,1.860);
\node[num] at (0.310,1.550) {0};
\path[fill=blue!33] (0.620,1.240) rectangle (1.240,1.860);
\draw[white, line width=.15pt] (0.620,1.240) rectangle (1.240,1.860);
\node[num] at (0.930,1.550) {4};
\path[fill=blue!25] (1.240,1.240) rectangle (1.860,1.860);
\draw[white, line width=.15pt] (1.240,1.240) rectangle (1.860,1.860);
\node[num] at (1.550,1.550) {2};
\path[fill=blue!25] (1.860,1.240) rectangle (2.480,1.860);
\draw[white, line width=.15pt] (1.860,1.240) rectangle (2.480,1.860);
\node[num] at (2.170,1.550) {2};
\path[fill=blue!18] (2.480,1.240) rectangle (3.100,1.860);
\draw[white, line width=.15pt] (2.480,1.240) rectangle (3.100,1.860);
\node[num] at (2.790,1.550) {1};
\path[fill=blue!33] (3.100,1.240) rectangle (3.720,1.860);
\draw[white, line width=.15pt] (3.100,1.240) rectangle (3.720,1.860);
\node[num] at (3.410,1.550) {4};
\path[fill=blue!39] (3.720,1.240) rectangle (4.340,1.860);
\draw[white, line width=.15pt] (3.720,1.240) rectangle (4.340,1.860);
\node[numdark] at (4.030,1.550) {6};
\node[axislab, anchor=east] at (-.08,2.170) {Japan};
\path[fill=blue!36] (0.000,1.860) rectangle (0.620,2.480);
\draw[white, line width=.15pt] (0.000,1.860) rectangle (0.620,2.480);
\node[numdark] at (0.310,2.170) {5};
\path[fill=blue!29] (0.620,1.860) rectangle (1.240,2.480);
\draw[white, line width=.15pt] (0.620,1.860) rectangle (1.240,2.480);
\node[num] at (0.930,2.170) {3};
\path[fill=blue!29] (1.240,1.860) rectangle (1.860,2.480);
\draw[white, line width=.15pt] (1.240,1.860) rectangle (1.860,2.480);
\node[num] at (1.550,2.170) {3};
\path[fill=blue!25] (1.860,1.860) rectangle (2.480,2.480);
\draw[white, line width=.15pt] (1.860,1.860) rectangle (2.480,2.480);
\node[num] at (2.170,2.170) {2};
\path[fill=blue!29] (2.480,1.860) rectangle (3.100,2.480);
\draw[white, line width=.15pt] (2.480,1.860) rectangle (3.100,2.480);
\node[num] at (2.790,2.170) {3};
\path[fill=blue!25] (3.100,1.860) rectangle (3.720,2.480);
\draw[white, line width=.15pt] (3.100,1.860) rectangle (3.720,2.480);
\node[num] at (3.410,2.170) {2};
\path[fill=blue!47] (3.720,1.860) rectangle (4.340,2.480);
\draw[white, line width=.15pt] (3.720,1.860) rectangle (4.340,2.480);
\node[numdark] at (4.030,2.170) {10};
\node[axislab, anchor=east] at (-.08,2.790) {Turkey};
\path[fill=blue!39] (0.000,2.480) rectangle (0.620,3.100);
\draw[white, line width=.15pt] (0.000,2.480) rectangle (0.620,3.100);
\node[numdark] at (0.310,2.790) {6};
\path[fill=blue!36] (0.620,2.480) rectangle (1.240,3.100);
\draw[white, line width=.15pt] (0.620,2.480) rectangle (1.240,3.100);
\node[numdark] at (0.930,2.790) {5};
\path[fill=blue!25] (1.240,2.480) rectangle (1.860,3.100);
\draw[white, line width=.15pt] (1.240,2.480) rectangle (1.860,3.100);
\node[num] at (1.550,2.790) {2};
\path[fill=blue!29] (1.860,2.480) rectangle (2.480,3.100);
\draw[white, line width=.15pt] (1.860,2.480) rectangle (2.480,3.100);
\node[num] at (2.170,2.790) {3};
\path[fill=blue!18] (2.480,2.480) rectangle (3.100,3.100);
\draw[white, line width=.15pt] (2.480,2.480) rectangle (3.100,3.100);
\node[num] at (2.790,2.790) {1};
\path[fill=blue!18] (3.100,2.480) rectangle (3.720,3.100);
\draw[white, line width=.15pt] (3.100,2.480) rectangle (3.720,3.100);
\node[num] at (3.410,2.790) {1};
\path[fill=blue!47] (3.720,2.480) rectangle (4.340,3.100);
\draw[white, line width=.15pt] (3.720,2.480) rectangle (4.340,3.100);
\node[numdark] at (4.030,2.790) {10};
\node[axislab, anchor=east] at (-.08,3.410) {U.S.};
\path[fill=blue!6] (0.000,3.100) rectangle (0.620,3.720);
\draw[white, line width=.15pt] (0.000,3.100) rectangle (0.620,3.720);
\node[num] at (0.310,3.410) {0};
\path[fill=blue!36] (0.620,3.100) rectangle (1.240,3.720);
\draw[white, line width=.15pt] (0.620,3.100) rectangle (1.240,3.720);
\node[numdark] at (0.930,3.410) {5};
\path[fill=blue!18] (1.240,3.100) rectangle (1.860,3.720);
\draw[white, line width=.15pt] (1.240,3.100) rectangle (1.860,3.720);
\node[num] at (1.550,3.410) {1};
\path[fill=blue!25] (1.860,3.100) rectangle (2.480,3.720);
\draw[white, line width=.15pt] (1.860,3.100) rectangle (2.480,3.720);
\node[num] at (2.170,3.410) {2};
\path[fill=blue!33] (2.480,3.100) rectangle (3.100,3.720);
\draw[white, line width=.15pt] (2.480,3.100) rectangle (3.100,3.720);
\node[num] at (2.790,3.410) {4};
\path[fill=blue!29] (3.100,3.100) rectangle (3.720,3.720);
\draw[white, line width=.15pt] (3.100,3.100) rectangle (3.720,3.720);
\node[num] at (3.410,3.410) {3};
\path[fill=blue!56] (3.720,3.100) rectangle (4.340,3.720);
\draw[white, line width=.15pt] (3.720,3.100) rectangle (4.340,3.720);
\node[numdark] at (4.030,3.410) {18};
\node[axislab, anchor=east] at (-.08,4.030) {France};
\path[fill=blue!36] (0.000,3.720) rectangle (0.620,4.340);
\draw[white, line width=.15pt] (0.000,3.720) rectangle (0.620,4.340);
\node[numdark] at (0.310,4.030) {5};
\path[fill=blue!29] (0.620,3.720) rectangle (1.240,4.340);
\draw[white, line width=.15pt] (0.620,3.720) rectangle (1.240,4.340);
\node[num] at (0.930,4.030) {3};
\path[fill=blue!29] (1.240,3.720) rectangle (1.860,4.340);
\draw[white, line width=.15pt] (1.240,3.720) rectangle (1.860,4.340);
\node[num] at (1.550,4.030) {3};
\path[fill=blue!39] (1.860,3.720) rectangle (2.480,4.340);
\draw[white, line width=.15pt] (1.860,3.720) rectangle (2.480,4.340);
\node[numdark] at (2.170,4.030) {6};
\path[fill=blue!6] (2.480,3.720) rectangle (3.100,4.340);
\draw[white, line width=.15pt] (2.480,3.720) rectangle (3.100,4.340);
\node[num] at (2.790,4.030) {0};
\path[fill=blue!25] (3.100,3.720) rectangle (3.720,4.340);
\draw[white, line width=.15pt] (3.100,3.720) rectangle (3.720,4.340);
\node[num] at (3.410,4.030) {2};
\path[fill=blue!58] (3.720,3.720) rectangle (4.340,4.340);
\draw[white, line width=.15pt] (3.720,3.720) rectangle (4.340,4.340);
\node[numdark] at (4.030,4.030) {20};
\node[axislab, rotate=45, anchor=west] at (0.310,4.420) {U.S.};
\node[axislab, rotate=45, anchor=west] at (0.930,4.420) {Italy};
\node[axislab, rotate=45, anchor=west] at (1.550,4.420) {Russia};
\node[axislab, rotate=45, anchor=west] at (2.170,4.420) {Brazil};
\node[axislab, rotate=45, anchor=west] at (2.790,4.420) {France};
\node[axislab, rotate=45, anchor=west] at (3.410,4.420) {Hungary};
\node[axislab, rotate=45, anchor=west] at (4.030,4.420) {Other};
\node[axislab, anchor=south] at (2.170,5.160) {surname-source country};
\node[axislab, rotate=90, anchor=south] at (-1.02,2.170) {given-source country};
\end{tikzpicture}
\caption{Source-country pair distribution for the 300 accepted operationally unknown names.}
\label{fig:wikidata-accepted-country-pairs}
\end{figure}

\subsection{Name Validation Yield}
\label{sec:resource-validation-yield}

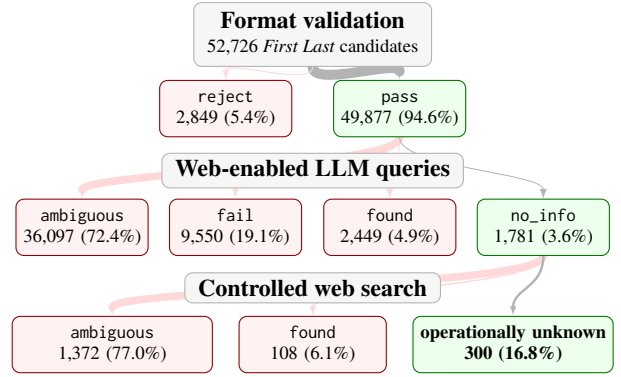
\begin{figure}[h]
\centering 
\begin{tikzpicture}[x=1cm,y=1cm,
  neutralbox/.style={draw=gray!60, fill=gray!7, rounded corners=3pt,
    align=center, minimum height=0.54cm,
    inner xsep=2.8pt, inner ysep=2.6pt, font=\scriptsize},
  passbox/.style={draw=green!45!black, fill=green!7, rounded corners=3pt,
    align=center, minimum height=0.54cm,
    inner xsep=2.8pt, inner ysep=2.6pt, font=\scriptsize},
  rejectbox/.style={draw=red!55!black, fill=red!5, rounded corners=3pt,
    align=center, minimum height=0.54cm,
    inner xsep=2.8pt, inner ysep=2.6pt, font=\scriptsize},
  mainbox/.style={#1box, text width=1.55cm},
  webbox/.style={#1box, text width=1.55cm},
  searchbox/.style={#1box, text width=1.75cm},
  searchwide/.style={#1box, text width=2.45cm},
  flow/.style={-{Latex[length=1.25mm,width=0.9mm]}, line cap=round,
    rounded corners=2pt},
  passflow/.style={flow, draw=gray!60},
  rejectflow/.style={flow, draw=red!15},
  title/.style={font=\small\bfseries, align=center, text=black,
    fill=gray!7, draw=gray!60, rounded corners=3pt,
    inner xsep=5pt, inner ysep=2.5pt},
]
\node[title] (format_title) at (0,0)
  {\begin{tabular}{@{}c@{}}
  Format validation\\[-1pt]
  {\normalfont\scriptsize 52,726 \emph{First Last} candidates}
  \end{tabular}};

\node[mainbox=pass] (local_pass) at (1.15,-0.98)
  {\begin{tabular}{@{}c@{}}
  \texttt{pass}\\
  49,877 (94.6\%)
  \end{tabular}};

\node[mainbox=reject] (local_reject) at (-1.15,-0.98)
  {\begin{tabular}{@{}c@{}}
  \texttt{reject}\\
  2,849 (5.4\%)
  \end{tabular}};

\node[title] (web_title) at (0,-1.8) {Web-enabled LLM queries};

\node[webbox=reject] (az_ambiguous) at (-3.05,-2.55)
  {\begin{tabular}{@{}c@{}}
  \texttt{ambiguous}\\
  36,097 (72.4\%)
  \end{tabular}};

\node[webbox=reject] (az_fail) at (-1.02,-2.55)
  {\begin{tabular}{@{}c@{}}
  \texttt{fail}\\
  9,550 (19.1\%)
  \end{tabular}};

\node[webbox=reject] (az_found) at (1.02,-2.55)
  {\begin{tabular}{@{}c@{}}
  \texttt{found}\\
  2,449 (4.9\%)
  \end{tabular}};

\node[webbox=pass] (az_no_info) at (3.05,-2.55)
  {\begin{tabular}{@{}c@{}}
  \texttt{no\_info}\\
  1,781 (3.6\%)
  \end{tabular}};

\node[title] (search_title) at (0,-3.35) {Controlled web search};

\node[searchwide=reject] (sp_ambig) at (-2.65,-4.1)
  {\begin{tabular}{@{}c@{}}
  \texttt{ambiguous}\\
  1,372 (77.0\%)
  \end{tabular}};

\node[searchbox=reject] (sp_found) at (0,-4.1)
  {\begin{tabular}{@{}c@{}}
  \texttt{found}\\
  108 (6.1\%)
  \end{tabular}};

\node[searchwide=pass] (sp_accept) at (2.65,-4.1)
  {\begin{tabular}{@{}c@{}}
  \textbf{operationally unknown}\\
  \textbf{300 (16.8\%)}
  \end{tabular}};

\begin{pgfonlayer}{background}
\draw[passflow, line width=3.76pt]
  (format_title.south) .. controls +(0,-0.28) and +(0,0.28) ..
  (local_pass.north);

\draw[rejectflow, line width=0.40pt]
  (format_title.south) .. controls +(0,-0.28) and +(0,0.28) ..
  (local_reject.north);

\draw[rejectflow, line width=2.92pt]
  (local_pass.south) .. controls +(0,-0.40) and +(0,0.34) ..
  (az_ambiguous.north);

\draw[rejectflow, line width=0.92pt]
  (local_pass.south) .. controls +(0,-0.40) and +(0,0.34) ..
  (az_fail.north);

\draw[rejectflow, line width=0.36pt]
  (local_pass.south) .. controls +(0,-0.40) and +(0,0.34) ..
  (az_found.north);

\draw[passflow, line width=0.32pt]
  (local_pass.south) .. controls +(0,-0.40) and +(0,0.34) ..
  (az_no_info.north);

\draw[rejectflow, line width=2.87pt]
  (az_no_info.south) .. controls +(0,-0.42) and +(0,0.36) ..
  (sp_ambig.north);

\draw[rejectflow, line width=0.54pt]
  (az_no_info.south) .. controls +(0,-0.42) and +(0,0.36) ..
  (sp_found.north);

\draw[passflow, line width=0.92pt]
  (az_no_info.south) .. controls +(0,-0.42) and +(0,0.36) ..
  (sp_accept.north);
\end{pgfonlayer}

\end{tikzpicture}
\caption{
Validation yield for our showcase run of \textsc{pun}.
}
\label{fig:pun-validation-yield}
\end{figure}

Figure~\ref{fig:pun-validation-yield} summarises the validation yield from 52,726 generated candidates. Local checks remove 2,849 malformed or out-of-scope strings before web-facing validation. Among the remaining candidates, the web-enabled LLM screen is dominated by ambiguity: 36,097 candidates are linked to nearby names, variants, famous-name attractors, or other referents, while 2,449 are tied to exact-person evidence. A further 9,550 produce failed outputs, largely because of policy-like refusals. Only 1,781 candidates receive a \texttt{no\_info} label and proceed to controlled search.

Controlled search rejects most of these web-LLM \texttt{no\_info} candidates. It finds 108 exact-person traces and 1,372 ambiguity signals, leaving 300 accepted names. 
The accepted names therefore make up 0.6\% of the generated candidates (roughly one name per 176 candidates). Figure~\ref{fig:wikidata-accepted-country-pairs} shows the source-country pair distribution for these names.

\begin{figure}[h]
  \centering
  \includegraphics[width=1.0\columnwidth]{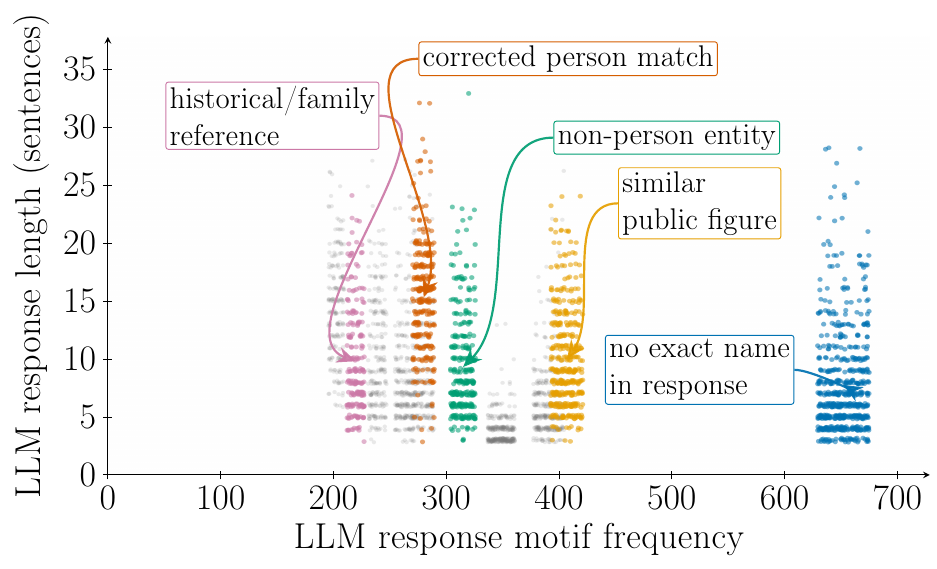}
  \caption{LLM response motifs for \texttt{ambiguous}.}
  \label{fig:llm-ambiguous-clusters}
\end{figure}

\begin{figure}[h]
  \centering
  \includegraphics[width=0.9\columnwidth]{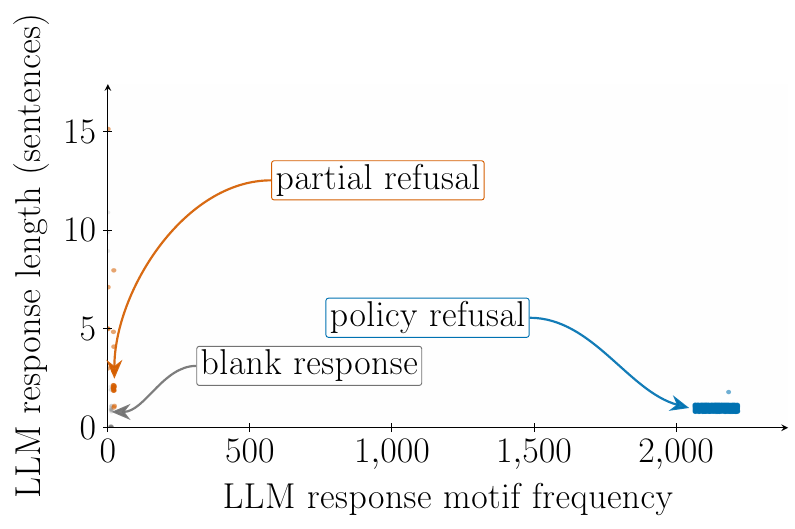}
  \caption{LLM response motifs for \texttt{fail}.}
  \label{fig:llm-fail-clusters}
\end{figure}

The rejected cases are informative because they show why exact quoted search alone is insufficient. Figure~\ref{fig:llm-ambiguous-clusters} groups common web-LLM ambiguity motifs, including similar names, partial-token matches, reordered forms, and nearby public referents. Figure~\ref{fig:llm-fail-clusters} groups failed responses, such as policy-like refusals. These patterns motivate the controlled-search ablations reported in Section~\ref{sec:results-ablations}.

\section{Experimental Setup}
\label{sec:experimental-setup}

\subsection{Protocol Reproducibility}
\label{sec:setup-reproducibility}

The validation stack contains two hosted model components: the web-enabled LLM (\texttt{gpt-4.1-mini}) that retrieves or synthesises evidence, and the adjudicator LLM (\texttt{gpt-5.2}) that maps the response to a discrete label. Even with fixed prompts and zero temperature, these are not fully deterministic. Following common practice in recent LLM evaluations, we therefore estimate stochastic variation with three repeated reruns on pre-specified cohorts (20\% of the 52,726 candidates). We use the same model versions, prompts, hyperparameters, and database snapshot, and report agreement over the paper-facing labels \texttt{found}, \texttt{ambiguous}, \texttt{no\_info}, and \texttt{fail}.

\subsection{Protocol Ablations}
\label{sec:protocol-ablations}

We ablate only the controlled-search validator, using the 1,781 candidates that passed web-LLM screening with \texttt{no\_info}. Each ablation removes one search guardrail and reports how many additional candidates would be accepted relative to the full validator. Generation-policy ablations are reported in Appendix~\ref{sec:appendix-ablations}.

\subsection{Name-Likeness}
\label{sec:setup-downstream}

We evaluate whether accepted \textsc{pun} names are technically name-like under two string-level diagnostics: character-level predictability and tokenizer fragmentation. For character-level predictability, we train an LSTM language model on ParaNames, a multilingual corpus of 140 million names across 400+ languages \citep{saleva-lignos-2024-paranames}, and score each evaluation string by bits per character (BPC). Lower BPC indicates that a string is more predictable under the name model. We compare accepted operationally unknown names against public-person controls, famous-name perturbation controls, and three non-name control sets.

For tokenizer fragmentation, we use the fact that subword tokenizers tend to keep frequent strings or fragments as larger units and split rarer forms into more pieces. For each name \(n\) and tokenizer \(T\), we compute \( |T(n)|/|n| \), and use the median across 14 tokenizers\footnote{
OpenAI \texttt{r50k\_base}, \texttt{p50k\_base}, \texttt{cl100k\_base}, and \texttt{o200k\_base}; GPT-2, RoBERTa, OPT, BERT, multilingual BERT, XLM-R, mT5, mBART-50, Qwen2.5, and Mistral.} as the name's fragmentation score. Appendix~\ref{sec:appendix-lstm-name-likeness} gives additional LSTM details, and Appendix~\ref{sec:appendix-tokenizer-footprint} gives the tokenizer-footprint diagnostics.

\subsection{Human Agreement}
\label{sec:setup-human-agreement}

We conducted a Prolific study with $n$ = 204 participants from 27 self-reported language backgrounds and 31 countries of residence. The study asked whether people make the same distinctions as the protocol. Participants first judged whether a string looks like a plausible full name, covering (i) accepted unknown names, (ii) real-person names that should be findable, and (iii) strings that should clearly not be names. Then, they inspected evidence by opening prepared Google, Bing, and DuckDuckGo queries in their own browsers, as well as by using web-enabled ChatGPT, Gemini, or Perplexity to search for selected unknown and ambiguous names, and relabelling saved web-LLM outputs. The study covers 495 unique person-name strings: 300 released unknown names, 108 controlled-search positives, 12 controlled-search ambiguous cases added for live-chatbot lookup, and 75 additional names unique to saved web-LLM outputs for classifier-agreement labelling. This allows us to separate name form plausibility from subjective human judgement, automated web-LLM screening from human-driven chatbot lookups, controlled-search agreement, and human--LLM classifier-label agreement. Detailed study design and setup are in Appendix~\ref{sec:appendix-human-agreement}.

\section{Results}
\label{sec:results}

\subsection{Protocol Reproducibility}
\label{sec:results-reproducibility}

End-to-end agreement is 75.0\% with very small between-run variation (standard deviation 0.1 percentage points), while classifier-only agreement on the saved original web responses is 98.2\%. This indicates that most instability comes from rerunning the web-enabled model and its search context, not from the discrete classifier.

The \texttt{no\_info} label is substantially less stable than the classifier itself (31.6\% end-to-end agreement versus 92.4\% classifier-only agreement in the ten-replicate \texttt{no\_info} rerun). In ten full reruns of the source \texttt{no\_info} cohort, only around 31\% of labels remain \texttt{no\_info} in any single rerun. Most switches are to \texttt{ambiguous} (57.7\%), while switches to \texttt{found} are rare (1.4\%). This supports the design choice that web-LLM \texttt{no\_info} is only a triage signal, where acceptance as operationally unknown requires the subsequent controlled-search stage. If we impose the stricter name-level criterion that a released name must remain \texttt{no\_info} in every web-LLM rerun, 49 of the 300 released names survive three reruns and 27 survive ten reruns. We refer to these stricter subsets as \texttt{stable\_no\_info\_3} and \texttt{stable\_no\_info\_10}.

\subsection{Protocol Ablations}
\label{sec:results-ablations}

Figure~\ref{fig:pun-validation-yield} shows the main validation attrition: only 300 of the 1,781 web-LLM \texttt{no\_info} candidates survive controlled search. Table~\ref{tab:controlled-search-ablation} asks which controlled-search guardrails account for this reduction. Exact quoted search alone would accept 1,115 names, i.e., 815 additional names rejected by the full validator. Removing reordered, ASCII-folded, vowel, and phonetic variants would accept 942 names, making variant search the largest search-side guardrail.

\begin{table}[h]
\centering
\scriptsize
\setlength{\tabcolsep}{4pt}
\renewcommand{\arraystretch}{0.92}
\begin{tabular}{lrr}
\toprule
Validator & Accepted & $\Delta$ \\
\midrule
Full validator & 300 & -- \\
Exact quoted only & 1,115 & +815 \\
No fuzzy/reordered variants & 942 & +642 \\
No unquoted queries & 356 & +56 \\
\bottomrule
\end{tabular}
\caption{Controlled-search ablations on the 1,781 web-LLM \texttt{no\_info} candidates ($\Delta$ relative to full validator).}
\label{tab:controlled-search-ablation}
\end{table}

\begin{figure}[t]
  \centering
  \includegraphics[width=1.0\columnwidth]{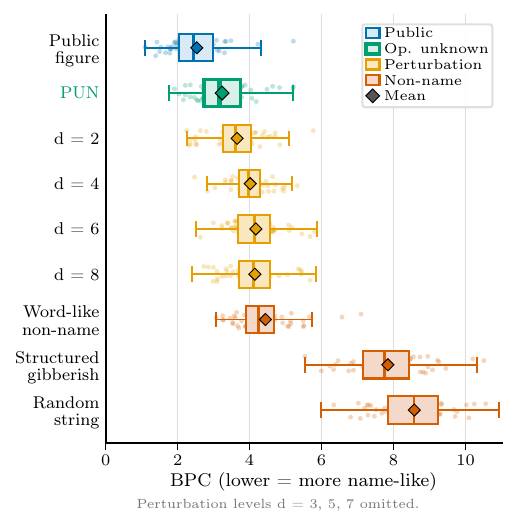}
  \caption{LSTM name-likeness by group, measured as bits per character (BPC; lower = more name-like).}
  \label{fig:lstm-boxplots}
\end{figure}

\subsection{Name-Likeness}
\label{sec:results-plausibility}

\begin{figure}[t]
  \centering
  \includegraphics[width=1.0\columnwidth]{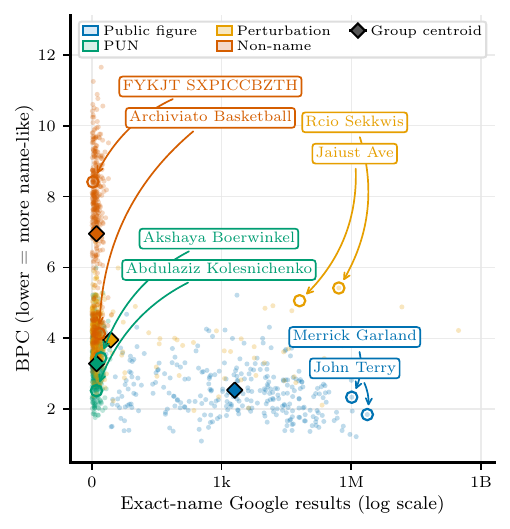}
  \caption{Web footprint versus LSTM name-likeness.}
  \label{fig:lstm-scatter}
\end{figure}

Our LSTM and tokenizer experiments show that accepted \textsc{pun} names are more name-like than famous-name perturbations and non-name controls, but less predictable than public-person names. In Figure~\ref{fig:lstm-boxplots}, lower bits per character (BPC) means that a string is more predictable under the ParaNames-trained name model. Accepted \textsc{pun} names have median BPC 3.16, above public-person names (2.45) but below all displayed famous-name perturbations and non-name controls, including word-like non-names (4.24), structured gibberish (7.75), and random strings (8.57). Figure~\ref{fig:lstm-scatter} visualises how public-person names are highly name-like but have a large web footprint, while random strings have little footprint but are not name-like. Accepted \textsc{pun} names instead cluster close to the ideal lower-left region with few exact-name Google results and low BPC.

Tokenizer fragmentation supports the same interpretation at the level of name parts. Accepted \textsc{pun} names have median fragmentation 0.36 pieces per character, close to public-person names and word-like non-names (both 0.33), and far below random strings (0.59). Fragmentation correlates with LSTM BPC across all scored rows (\(r=0.74\), Spearman \(\rho=0.63\)), but only weakly with exact-name search prevalence (Spearman \(\rho=-0.16\)).  Appendix Figures~\ref{fig:tokenizer-fragmentation-boxplots} and~\ref{fig:tokenizer-fragmentation-scatter} report the tokenizer plots.

\subsection{Human Agreement}
\label{sec:results-human-agreement}

The human agreement audit tests whether the protocol's evidential distinctions are recoverable by non-expert participants. We separate four questions: whether accepted names look like plausible full names; whether participants can find person evidence for names accepted as operationally unknown; whether they can recover evidence for names rejected by controlled search as \texttt{found}; and whether they assign the same labels as the automated classifier when shown saved web-LLM outputs. Participants were given definitions of \texttt{found}, \texttt{ambiguous}, \texttt{no\_info}, and \texttt{fail}, and had to pass a comprehension check before beginning the study.

\paragraph{Name plausibility.} Participants judged the 300 released unknown names as plausible full names in 63.0\% of ratings. By comparison, names rejected by \textsc{pun}, were judged plausible in 77.2\% of ratings. The non-name controls were rarely rated as plausible: 9.8\% for word-like no-names, 0.0\% for structured gibberish, and 0.5\% for random character strings. This shows that operational plausibility, as validated via LSTM, and human-perceived plausibility are related but distinct properties.

\paragraph{Controlled-search agreement.} Participants searched the 300 released unknown names using Google, Bing, and DuckDuckGo. They selected \texttt{no\_info} in 75.8\% of judgments and \texttt{found} in 3.0\%. Most remaining disagreement was \texttt{ambiguous} (17.2\%), typically reflecting nearby names, partial-token matches, spelling variants, or search-provider corrections. The stricter \texttt{stable\_no\_info\_10} subset behaves more strongly under the same audit: participants selected \texttt{no\_info} in 84.0\% of judgments, \texttt{ambiguous} in 12.3\%, and \texttt{found} in 2.5\%. Thus, human search mostly supports the released verdicts and confirms that ambiguity is the main boundary case.

\paragraph{Web-enabled chatbots.} When participants used an assigned web-enabled chatbot to search for the 300 operationally unknown names, they labelled the chatbot response as \texttt{no\_info} in 82.8\% of cases, \texttt{ambiguous} in 12.1\%, \texttt{found} in 2.0\%, and \texttt{fail} in 3.0\%. The \texttt{found} rate was concentrated in ChatGPT responses (5.4\%), with 0.0\% for Gemini and 0.0\% for Perplexity. For names where the \textsc{pun} web-enabled LLM returned \texttt{no\_info}, participants using chatbots also usually failed to find evidence. The \texttt{no\_info} rate was 88.9\% with ChatGPT, 100.0\% with Gemini, and 88.2\% with Perplexity.

Finally, the saved-response labelling task checks whether participants can recognise exact-person evidence when it is present in a web-LLM output. They separate exact-person evidence from all other outcomes in 88.7\% of cases. Fine-grained four-way agreement is lower, especially for \texttt{ambiguous} and \texttt{fail}, because these labels require distinguishing near-match evidence, absence statements, and unusable responses. Overall, the human audit supports the main protocol decision: accepted names are generally perceived as names and yield hardly any recoverable person evidence, with disagreement being concentrated in ambiguous boundary cases that our reproducibility tests surfaced too.

%

%

\section{Discussion}
\label{sec:discussion}

\textsc{pun} is best understood as a measurement device, not as a catalogue of non-existent people. Our experiments confirm that operationally unknown names remain name-like, and human study participants are unable to recover person evidence in 97\% of cases. At the same time, the claim remains operational, as verdicts can change based on demographics and social associations.

Our protocol enables controlled follow-on studies. For example, accepted names can be paired with synthetic facts to test whether models reproduce, distort, protect, or forget introduced person records. They can serve as negative controls in privacy audits, since a method that extracts claims for operationally unknown names may not be measuring leakage alone. They can support name-bias studies with fewer confounds from known individuals, and machine-unlearning benchmarks where the inserted records and deletion targets are known even when original pretraining data are opaque. Revalidating the same unknown names over time can also measure how search systems and retrieval-augmented models begin attaching information to a previously unsupported full-name string.

\section{Conclusion}
\label{sec:conclusion}

We introduced \textsc{pun}, a protocol for constructing plausible person-name prompts that are operationally unknown under a documented validation run. We demonstrated that person names should not be treated as interchangeable prompt strings. Their evidential status is an experimental variable. \textsc{pun} supplies one missing baseline for measuring what LLMs do when a prompt has person-name form but lacks validated public full-name evidence. Future work should extend the protocol beyond Latin-script two-component names, quantify provider and locale sensitivity, and integrate operationally unknown names into downstream benchmarks.

\newpage

\section*{Limitations}
\label{sec:limitations}

Operational unknownness is provider-, configuration-, locale-, and time-dependent. Search engines may miss evidence, change indexes, alter spell correction, or report unstable result counts. Hosted LLMs used for screening and adjudication may also change over time, even when prompts and decoding settings are fixed. The released run should therefore be read as a documented validation snapshot rather than as a permanent property of the accepted strings.

The present resource is restricted to Latin-script two-component \emph{First Last} strings. It does not cover non-Latin scripts, transliteration, patronymics, mononyms, multi-part surnames, name particles, culturally variable name order, or many other naming practices. The Wikidata-derived source pools are not demographically representative, and the source-country key is only a sampling and audit proxy, not a claim about nationality, ethnicity, language, or cultural identity.

The 300-name seed set is intentionally small and conservative. It is intended as a showcase resource and evaluation control, not as a comprehensive catalogue of unknown names. Before high-stakes or time-sensitive use, users should revalidate names under a newly documented \(PUN_t\). Absence of indexed full-name evidence under \(PUN_t\) is not evidence that no person has the name, that no private or offline record contains it, or that no model has encountered the string during training.

Finally, the protocol controls full-name evidence and ambiguity, but does not remove all social meaning from names. Given-name and surname components may still carry associations with language, region, gender, religion, class, ethnicity, or other perceived attributes. Evaluations using \textsc{pun} should therefore interpret model behaviour as conditioned on plausible name strings with controlled full-name evidence, not demographically neutral identifiers.

\section*{Ethical Considerations}
\label{sec:ethics}

This work constructs realistic-looking person-name prompts for evaluation. Such names may coincide with real but non-indexed individuals, or with people whose public evidence was not found by the validation run. In addition, the public-person comparison controls intentionally contain names of real public figures sampled from Wikidata. The resource must therefore not be used to claim that accepted names prove non-existence, to fabricate biographies, to impersonate real people, or to target people with the same or similar names. Public-person controls should likewise be used only as aggregate controls, not for individual-level claims about the named persons. The intended use of the resource is aggregate evaluation of LLM behaviour, including factuality, abstention, privacy-related behaviour, and name-conditioned conflation.


The protocol uses public web evidence and Wikidata-derived name components. Where raw search traces contain personal information, we minimise disclosure by releasing hashes, labels, query metadata, and summary evidence rather than unnecessary personal snippets. Repository maintainers should avoid redistributing sensitive personal details that are not needed for reproducibility.

We conducted a human agreement audit with Prolific participants. Participants consented, used a desktop or laptop browser, and completed comprehension and attention checks. The study collected task responses and platform metadata for assignment, exclusion, and aggregate reporting. Participants were paid USD 8.09/h for a median completion time of 8.2 minutes. No participant-level identifying information is released.

\paragraph{LLM-based Tools.}
We used LLM-based tools in a limited way during manuscript preparation and implementation. GitHub Copilot was used for code completion and minor refactoring, while GPT~5.5 was used to suggest alternative phrasings, and polish Matplotlib visualisations and \LaTeX{} formatting, e.g., table layouts in the appendix sections. All algorithmic design decisions, experimental implementation and execution, data analysis, and substantive writing were carried out by the authors, and we verified all AI-assisted edits for correctness.

\bibliography{custom}

\appendix

\section{Appendix}
\label{sec:appendix}

This appendix provides supplementary details for the main analyses: candidate generation
(Appendix~\ref{sec:appendix-candidate-generation}), protocol implementation
(Appendix~\ref{sec:appendix-protocol-details}), comparison-name construction
(Appendix~\ref{sec:appendix-comparison-names}), ablations
(Appendix~\ref{sec:appendix-ablations}), name-likeness diagnostics
(Appendices~\ref{sec:appendix-lstm-name-likeness} and
\ref{sec:appendix-tokenizer-footprint}), and the human agreement study
(Appendix~\ref{sec:appendix-human-agreement}).

\subsection{Name Candidate Generation Details}
\label{sec:appendix-candidate-generation}

This appendix records the construction choices behind Section~\ref{sec:resource-candidate-generation}. We store the source entity, the original label, the component role, and a source-country key derived from the source person's recorded birthplace. The source-country key is used only for sampling and audit. It is not a claim about the generated name or about the source person's nationality, ethnicity, language, or demographic status.

\paragraph{Coverage and script mix.}
The May 2026 Wikidata dump yielded 13.5M human entities. The selected labels are already heavily Latin-script, and the mapped source pool remains so after name parsing and birthplace-country mapping. Table~\ref{tab:wikidata-script-distribution-compact} summarises this skew. These counts are extraction diagnostics, not demographic estimates.

\begin{table}[h]
\centering
\small
\begin{tabular}{lrrrr}
\toprule
Script & Selected lbs & Pct. & Mapped lbs & Pct. \\
\midrule
Latin & 13,039,639 & 96.35 & 4,032,787 & 96.76 \\
Cyrillic & 238,416 & 1.76 & 111,872 & 2.68 \\
Han/CJK & 118,179 & 0.87 & 210 & 0.01 \\
Arabic & 40,538 & 0.30 & 8,048 & 0.19 \\
Other & 115,758 & 0.85 & 14,814 & 0.36 \\
\bottomrule
\end{tabular}
\caption{Compact script distribution for the Wikidata human labels retained by the pipeline. ``Selected lbs'' are the one-label-per-human extraction output. ``mapped lbs'' are the subset with a parsable name and mapped birthplace-country key.}
\label{tab:wikidata-script-distribution-compact}
\end{table}

\paragraph{Source-country-key skew.}
The mapped source pool contains 4,167,732 humans across 667 source-country keys. Before filtering, Germany and the United States are the largest keys, and the top ten keys account for 56.3\% of the mapped pool. After filtering, 40,715 eligible component slots remain across 26 country keys, with U.S., French, German, Italian, and Russian source pools accounting for 70.5\% combined. Table~\ref{tab:wikidata-country-skew-compact} gives a compact summary.

\begin{table}[h]
\centering
\small
\begin{tabular}{lrrrr}
\toprule
Source-country key & MSH & Pct. & ECS & Pct. \\
\midrule
Germany & 618,954 & 14.85 & 4,662 & 11.45 \\
United States & 433,649 & 10.40 & 8,869 & 21.78 \\
France & 219,734 & 5.27 & 8,177 & 20.08 \\
United Kingdom & 193,465 & 4.64 & 1,405 & 3.45 \\
Italy & 182,248 & 4.37 & 4,559 & 11.20 \\
Russia & 139,541 & 3.35 & 2,445 & 6.01 \\
Other keys & 2,380,341 & 57.12 & 10,598 & 26.03 \\
\bottomrule
\end{tabular}
\caption{Compact source-country-key skew before and after component filtering. MSH = mapped source humans; ECS = eligible component slots.}
\label{tab:wikidata-country-skew-compact}
\end{table}

\paragraph{Component filters.}
Tokenising the mapped source humans gives 356,558 distinct given-name components and 989,563 distinct surname components. Components are retained only if (a) they are sufficiently rare within their source-country pool, (b) appear in at least two source records, (c) have length at least six, (d) use Latin-script form, and (e) have non-empty native-language metadata. These filters reduce common-name collisions, parsing artefacts, very short tokens, and components with weak provenance. Table~\ref{tab:wikidata-filter-attrition} gives the cumulative attrition.

\begin{table}[h]
\centering
\small
\begin{tabular}{lrr}
\toprule
Cumulative filter & Given & Surname \\
\midrule
Raw pool & 356,558 & 989,563 \\
Rarity $p \leq 5{\times}10^{-5}$ & 259,192 & 756,262 \\
Pool occurrences $\geq 2$ & 64,692 & 221,573 \\
Length $\geq 6$ & 45,266 & 184,410 \\
Latin-script form & 38,841 & 182,578 \\
Native-language\\metadata present & 8,404 & 30,576 \\
\bottomrule
\end{tabular}
\caption{Cumulative reduction of distinct Wikidata-derived components under the release filters.}
\label{tab:wikidata-filter-attrition}
\end{table}

\paragraph{Candidate sampling and local validation.}
Candidate names are sampled by combining one eligible given-name component and one eligible surname component into a two-token \emph{First Last} string. We sample across source-country-key pairs rather than directly from raw component counts, and reject same-country recombinations. This prevents the largest Wikidata pools from dominating candidate generation and reduces the chance of reconstructing attested same-country full names.

Local validation is applied before any web-facing call. It rejects strings with invalid token count, unsupported script, malformed punctuation, duplicate form, titles, suffixes, excessive length, or components that fail the rarity, support, length, script, or provenance constraints. These checks define operational form plausibility for this paper run. They do not guarantee sociolinguistic naturalness, legal validity, demographic balance, or cross-cultural representativeness.

\subsection{Protocol Implementation Details}
\label{sec:appendix-protocol-details}

This appendix subsection records implementation details for the protocol summarised in Section~\ref{sec:protocol}. These details are part of the documented validation run \(PUN_t\) and should be fixed when reproducing or revalidating the released resource.

\paragraph{Wikidata extraction and component pools.}
We extract labels from Wikidata human entities and parse retained labels into candidate given-name and surname positions. For each component, we store the source entity, the original full-name label, the component role, and a source-country key derived from the source person's recorded birthplace. Components are retained only when they pass the local form constraints used by the release pipeline. In the paper run, candidates had to use Latin-script components, retain diacritics, have components of at least six characters, have no more than 60 characters in total, contain no titles or suffixes, and avoid unsupported punctuation. Components also had to have within-country-pool probability at most \(5\times10^{-5}\), appear in at least two source records, and have a non-empty Wikidata native-language entry.

\paragraph{Wikidata coverage, script mix, and country skew.}
Tables~\ref{tab:wikidata-script-distribution} and~\ref{tab:wikidata-country-skew} describe the Wikidata slice used by the pipeline. They do not describe every alias or every language-specific label stored in Wikidata. For each human entity, the extraction keeps one label under the paper run's language-priority policy. We then assign that retained label to a Unicode script bucket. These buckets are script diagnostics, not language, nationality, ethnicity, or demographic labels. The retained labels are mostly English-priority labels (86.2\%) or multilingual labels (7.8\%); the next largest label-language shares are Russian (1.1\%), Chinese (0.6\%), French (0.4\%), Japanese (0.2\%), German (0.2\%), Spanish (0.2\%), and Arabic (0.2\%).

\begin{table*}[t]
\centering
\small
\resizebox{\textwidth}{!}{%
\begin{tabular}{lrrrr}
\toprule
Script bucket & Selected Wikidata-human labels & Pct. & Mapped source-human labels & Pct. \\
\midrule
Latin & 13,039,639 & 96.35 & 4,032,787 & 96.76 \\
Cyrillic & 238,416 & 1.76 & 111,872 & 2.68 \\
Han/CJK & 118,179 & 0.87 & 210 & 0.01 \\
Arabic & 40,538 & 0.30 & 8,048 & 0.19 \\
Hangul/Korean & 20,700 & 0.15 & 42 & $<0.01$ \\
Hebrew & 14,348 & 0.11 & 4,097 & 0.10 \\
Devanagari & 6,633 & 0.05 & 140 & $<0.01$ \\
Japanese kana/mixed & 4,771 & 0.04 & 142 & $<0.01$ \\
Armenian & 4,766 & 0.04 & 2,856 & 0.07 \\
Thai & 4,510 & 0.03 & 314 & 0.01 \\
Greek & 3,827 & 0.03 & 1,394 & 0.03 \\
Bengali & 1,753 & 0.01 & 198 & $<0.01$ \\
Myanmar/Burmese & 1,035 & 0.01 & 174 & $<0.01$ \\
Georgian & 338 & $<0.01$ & 19 & $<0.01$ \\
Ethiopic/Amharic & 181 & $<0.01$ & 2 & $<0.01$ \\
Thaana & 40 & $<0.01$ & 5 & $<0.01$ \\
Tifinagh & 6 & $<0.01$ & 0 & 0.00 \\
Canadian Aboriginal syllabics/Inuktitut & 0 & 0.00 & 0 & 0.00 \\
Mongolian & 0 & 0.00 & 0 & 0.00 \\
Other or mixed script & 33,290 & 0.25 & 5,431 & 0.13 \\
\bottomrule
\end{tabular}
}
\caption{Script distribution of the Wikidata human labels retained by the pipeline. ``Selected labels'' are the one-label-per-human extraction output under the paper run's language-priority policy; ``mapped source-human labels'' are the subset with a parsable name and mapped birthplace-country key. Han/CJK and Japanese-kana buckets are Unicode script buckets, not language identifiers.}
\label{tab:wikidata-script-distribution}
\end{table*}

\begin{table}[t]
\centering
\small
\begin{tabular}{lrrrr}
\toprule
Source-country key & MSH & Pct. & ECS & Pct. \\
\midrule
Germany & 618,954 & 14.85 & 4,662 & 11.45 \\
United States & 433,649 & 10.40 & 8,869 & 21.78 \\
France & 219,734 & 5.27 & 8,177 & 20.08 \\
United Kingdom & 193,465 & 4.64 & 1,405 & 3.45 \\
Italy & 182,248 & 4.37 & 4,559 & 11.20 \\
Czech Republic & 156,921 & 3.77 & 561 & 1.38 \\
Spain & 140,670 & 3.38 & 1,200 & 2.95 \\
Japan & 140,229 & 3.36 & 1,744 & 4.28 \\
Russia & 139,541 & 3.35 & 2,445 & 6.01 \\
Poland & 121,743 & 2.92 & 989 & 2.43 \\
Austria & 74,694 & 1.79 & 124 & 0.30 \\
Ukraine & 70,073 & 1.68 & 364 & 0.89 \\
\bottomrule
\end{tabular}
\caption{Largest birthplace-derived source-country shares before and after release filters. The mapped source pool contains 4,167,732 humans across 667 country keys; the top ten account for 56.3\%. Eligible component slots are the remaining country-specific given-name and surname options after rarity, support, length, Latin-script, and native-language-metadata filters. The eligible set contains 40,715 slots across 26 country keys, with the top five accounting for 70.5\%. MSH = Mapped source humans. ECS = Eligible component slots.}
\label{tab:wikidata-country-skew}
\end{table}

\paragraph{Candidate sampling and local validation.}
Candidate names are sampled by combining one eligible given-name component and one eligible surname component into a two-token \emph{First Last} string. Local validation is applied before any web-facing call. It rejects strings with invalid token count, unsupported scripts, malformed punctuation, duplicate forms, titles, suffixes, excessive length, or components that fail the frequency and provenance constraints. These checks are intended to reduce parsing artefacts and extremely common name components, not to guarantee sociolinguistic naturalness.

\paragraph{Web-enabled LLM screening.}
For the paper run, locally valid candidates were queried with the prompt ``Who is \{name\}?''. We used a web-enabled model for broad evidence triage and a separate adjudicator model to map the response to a discrete label. The screen used \texttt{gpt-4.1-mini} with \texttt{web\_search\_preview} and \texttt{temperature: 0.0}; the adjudicator used \texttt{gpt-5.2} with \texttt{temperature: 0.0}. The adjudicator assigned one of four labels: \texttt{found}, \texttt{ambiguous}, \texttt{no\_info}, or \texttt{fail}. The full adjudicator prompt and label rubric are reported in Appendix~\ref{sec:appendix-adjudicator-rubric}. Only candidates labelled \texttt{no\_info} were passed to controlled search.

\paragraph{Controlled search queries.}
Controlled search was issued through Google Search via \texttt{serper.dev}. The verifier checked the exact full name and a bounded set of configured equivalents. Query forms included quoted and unquoted full-name queries, reversed-order variants, ASCII-folded variants for names with diacritics, exchanged-vowel variants, and phonetic variants. For example, a name such as \emph{Jos\'e N\'u\~nez} is also checked as \emph{Jose Nunez}, \emph{N\'u\~nez Jos\'e}, vowel-edited forms, and phonetic edits such as surname-final consonant substitutions. These variants should catch full-name evidence, search corrections, and near-name ambiguity that exact quoted search alone may miss.

\paragraph{Evidence and ambiguity rules.}
A candidate is rejected as \texttt{found} when a search-result title, snippet, URL, knowledge-panel field, or inspected page content contains the exact or configured full-name form in a person-like context. Person-like context is operationalised with a fixed list of biographical terms, person-oriented domains, and knowledge-panel type cues. A candidate is rejected as \texttt{ambiguous} when the evidence does not establish exact full-name person evidence but points to a nearby full-name form, spelling correction, reordered form, famous-person attractor, fictional character, organisation, place, work, product, or other unclear referent. Search corrections and configured variants that point to a different nearby full-name form are treated as ambiguity signals rather than as acceptance evidence.

\paragraph{Stored audit trace.}
For each candidate, the run stores provenance and validation metadata sufficient for later inspection: sampled components, source records, local-validation outcome, web-enabled LLM response hashes, adjudicated labels, controlled-search query forms, provider metadata, timestamps, evidence labels, verdict reasons, and final acceptance status. Because operational unknownness is time-bounded, later uses of the resource should either cite the original validation run or rerun the controlled-search stage under a newly documented \(PUN_t\).

\subsection{Adjudicator Prompt and Label Rubric}
\label{sec:appendix-adjudicator-rubric}

The web-enabled LLM screen uses a retrieval-style prompt, ``Who is \{name\}?''. The returned text is then classified by a separate adjudicator. The paper run used the following rubric, with JSON-only output:

\begin{quote}
\footnotesize
Choose exactly one label: \texttt{found}, \texttt{similar}, \texttt{attractor}, \texttt{no\_info}, or \texttt{fail}. Apply the decision order below and stop at the first match. \texttt{fail}: refusal or policy text, garbled output, error text, or insufficient information to classify. \texttt{found}: the exact full name is explicitly identified as a specific person with concrete biographical detail; multiple people with the exact name are not \texttt{found}, but an explicit alias or alternate-spelling equivalence to a specific person is \texttt{found}. \texttt{attractor}: the response treats the queried name as a misspelling, typo, or close variant of a known public person. \texttt{similar}: any other person or entity is mentioned that is not the exact full name, including partial-token matches, alternate spellings without explicit alias equivalence, fictional characters, organisations, places, works, products, and public, social, or genealogy records. \texttt{no\_info}: the response explicitly reports no information about the exact name and mentions no other people or entities; name meaning, etymology, or origin only also counts as \texttt{no\_info}. Exact-name matching is case- and diacritics-insensitive; all tokens must be present and token spelling must match, with surrounding titles allowed.
\end{quote}

For paper-facing verdicts, \texttt{found} maps to \texttt{web\_evidence\_found}; \texttt{similar} and \texttt{attractor} map to \texttt{ambiguous}; \texttt{fail} maps to \texttt{validation\_incomplete}; and \texttt{no\_info} is only a triage label. A name is accepted as \texttt{no\_indexed\_full\_name\_evidence\_found} only after controlled search also completes without exact-person evidence or ambiguity signals.

\subsection{Controlled-Search Query Inventory}
\label{sec:appendix-query-inventory}

Table~\ref{tab:controlled-search-query-inventory} gives the controlled-search query inventory used for the paper run. Query variants are generated deterministically from the normalised full-name string. The verifier records every executed query, its query type, provider metadata, search-result traces, and whether any trace is classified as \texttt{exact\_person\_trace}, \texttt{ambiguity\_signal}, \texttt{unrelated}, or \texttt{inconclusive}. Acceptance requires a clean web-LLM \texttt{no\_info} label and completion of all configured mandatory query types.

\begin{table}[h]
\centering
\scriptsize
\resizebox{\columnwidth}{!}{%
\begin{tabular}{lll}
\toprule
Query type & Form & Purpose \\
\midrule
\texttt{exact\_quoted} & ``\texttt{"First Last"}'' & exact full-name evidence \\
\texttt{unquoted} & \texttt{First Last} & ranked evidence and corrections \\
\texttt{reversed} & \texttt{"Last First"}; unquoted & order variants \\
\texttt{ascii\_fold} & diacritics removed & accent-insensitive evidence \\
\texttt{vowel\_variant} & one vowel edited & spelling-near variants \\
\texttt{phonetic\_variant} & one conservative edit & sound-near variants \\
\texttt{page\_content} & top-result page text & evidence beyond snippets \\
\bottomrule
\end{tabular}
}
\caption{Controlled-search query inventory. The paper run used Google Search via \texttt{serper.dev}, locale metadata stored with the run, a two-pass policy, page-content fetching for top organic results, and configured caps of 200 vowel variants, 50 phonetic variants, and 500 total variants per name.}
\label{tab:controlled-search-query-inventory}
\end{table}

The phonetic variant rules are deliberately conservative single-token substitutions: \texttt{ph}\(\leftrightarrow\)\texttt{f}, \texttt{ck}\(\leftrightarrow\)\texttt{k}, \texttt{c}\(\rightarrow\)\texttt{k}, \texttt{q}\(\rightarrow\)\texttt{k}, \texttt{z}\(\leftrightarrow\)\texttt{s}, \texttt{v}\(\leftrightarrow\)\texttt{w}, \texttt{y}\(\leftrightarrow\)\texttt{i}, \texttt{j}\(\leftrightarrow\)\texttt{g}, and \texttt{sh}\(\leftrightarrow\)\texttt{sch}. If an exact or configured full-name form appears in a person-like context, the verdict is \texttt{web\_evidence\_found}; if search hints, spelling corrections, variants, reordered forms, or nearby entities point to another plausible referent, the verdict is \texttt{ambiguous}; if mandatory queries fail or traces are inconclusive, the verdict is \texttt{validation\_incomplete}.

\subsection{Release Schema, Manifests, and Revalidation}
\label{sec:appendix-release-schema}

The public release separates accepted unknown names from comparison and rejected-control strata. Table~\ref{tab:release-schema-summary} summarises the paper-facing files and the key audit columns. Raw provider responses and search traces may contain personal data from indexed web pages; the release therefore prioritises names, labels, timestamps, query metadata, hashes, counts, verdict reasons, and compact trace summaries over unnecessary raw snippets.

\begin{table*}[t]
\centering
\small
\resizebox{\textwidth}{!}{%
\begin{tabular}{lll}
\toprule
Artifact & Rows in paper run & Key columns \\
\midrule
\textcolor{black}{\texttt{release/release\_unknown\_names.csv}} & 300 & \texttt{name}, \texttt{status}, \texttt{checked\_at}, \texttt{run\_id}, \texttt{validation\_stack\_hash}, \texttt{max\_query\_type}, \texttt{expected\_query\_count}, \texttt{executed\_query\_count} \\
\textcolor{black}{\texttt{release/comparison\_names.csv}} & 4,276 & accepted unknowns, public-person controls, proximity controls, stable-\texttt{no\_info} duplicate strata, comparison metadata \\
\textcolor{black}{\texttt{release/known\_person\_controls.csv}} & 300 & Wikidata QID, Wikipedia metadata, fame bin, exact-name prevalence, comparison level \\
\textcolor{black}{\texttt{release/ambiguous\_name\_controls.csv}} & 100 & rejected ambiguity controls with source status, decisive query type, and trace metadata \\
\textcolor{black}{\texttt{release/famous\_name\_perturbation\_controls.csv}} & 3,600 & source public person, generated variant, distance level, edit metadata, exact quoted prevalence \\
\textcolor{black}{\texttt{release/release\_manifest.json}} & \textcolor{black}{--} & \textcolor{black}{public manifest with release file paths, row counts, byte sizes, hashes, generation timestamp, and omitted-artifact notes} \\
\textcolor{black}{\texttt{release/reproducibility\_study\_20260521T095532Z\_noinfo3\_stable\_no\_info.csv}} & \textcolor{black}{49} & \textcolor{black}{accepted unknowns stable as \texttt{no\_info} across the three-replicate rerun panel} \\
\textcolor{black}{\texttt{release/reproducibility\_study\_20260521T095532Z\_noinfo10\_stable\_no\_info.csv}} & \textcolor{black}{27} & \textcolor{black}{accepted unknowns stable as \texttt{no\_info} across the ten-replicate rerun panel} \\
\textcolor{black}{\texttt{release/reproducibility\_study\_20260521T095532Z\_summary.json}} & \textcolor{black}{--} & \textcolor{black}{aggregate reproducibility-study summary for the rerun panels} \\
\bottomrule
\end{tabular}
}
\caption{Release-schema summary. Stable-\texttt{no\_info} rows duplicate accepted unknown names for analysis strata and are not additional unique names.}
\label{tab:release-schema-summary}
\end{table*}

For later use, the recommended procedure is to treat the released verdict as tied to its original \(PUN_t\), then run a new controlled-search validation before using the names in a fresh experiment. The implementation records a validation-stack hash from the local-filter, web-screen, and controlled-search configurations, so regenerated releases can be compared against the paper run. The linked repository uses:

\begin{quote}
\footnotesize
\texttt{python -m pun\_pipeline.run paper\_release \textbackslash}\\
\texttt{\ \ --dump /path/to/latest-all.json.bz2 \textbackslash}\\
\texttt{\ \ --target-unknowns 300 \textbackslash}\\
\texttt{\ \ --comparison-per-set 100 \textbackslash}\\
\texttt{\ \ --exact-perturbation-counts}
\end{quote}

Because publishing the accepted strings can itself make them indexed, later validation should record the release timestamp and ignore self-indexing traces only when they clearly originate from the dataset release rather than from independent person evidence.

\subsection{Reproducibility Details}
\label{sec:appendix-validation-protocol}

The reproducibility study was last run on 21 May 2026 using a source database snapshot with 49,880 latest web-screened candidates and 300 released operationally unknown names. For consistency with the release-yield analyses, the \texttt{no\_info}-only rows below are restricted to the 1,781 web-LLM \texttt{no\_info} names in the frozen 52,726-candidate release-yield cohort. The web-enabled model was \texttt{gpt-4.1-mini} with \texttt{web\_search\_preview}; the classifier was \texttt{gpt-5.2}. Both model calls used temperature 0.0. The runs reused the released prompts and classifier rubric, sample seed 13, and the same pre-specified sample across replicates. The study started at 2026-05-21T09:55:34Z and finished at 2026-05-21T13:52:28Z.

\begin{table*}[t]
\centering
\small
\resizebox{\textwidth}{!}{%
\begin{tabular}{llrrrrrrr}
\toprule
Rerun set & Cohort & Names & Label calls & \texttt{found} & \texttt{ambiguous} & \texttt{no\_info} & \texttt{fail} & Always \texttt{no\_info} \\
\midrule
3 reps & all source \texttt{no\_info} & 1,781 & 5,343 & 76 (1.4\%) & 3,115 (58.3\%) & 1,678 (31.4\%) & 474 (8.9\%) & 326 (18.3\%) \\
3 reps & released unknowns & 300 & 900 & 25 (2.8\%) & 546 (60.7\%) & 260 (28.9\%) & 69 (7.7\%) & 49 (16.3\%) \\
10 reps & all source \texttt{no\_info} & 1,781 & 17,810 & 244 (1.4\%) & 10,272 (57.7\%) & 5,635 (31.6\%) & 1,659 (9.3\%) & 189 (10.6\%) \\
10 reps & released unknowns & 300 & 3,000 & 74 (2.5\%) & 1,823 (60.8\%) & 857 (28.6\%) & 246 (8.2\%) & 27 (9.0\%) \\
\bottomrule
\end{tabular}
}
\caption{Detailed \texttt{no\_info}-only reproducibility outcomes. Counts are over rerun labels for names whose source web-screen label was \texttt{no\_info}. The \texttt{ambiguous} column combines rerun \texttt{similar} and \texttt{attractor} labels. ``Always \texttt{no\_info}'' is a name-level criterion: the name must receive \texttt{no\_info} in every replicate.}
\label{tab:reproducibility-noinfo-details}
\end{table*}

We ran three evaluations on a 20\% sample of the latest web-screened cohort (9,976 names), followed by three and ten full reruns of the source \texttt{no\_info} cohort (1,781 names per replicate). Table~\ref{tab:reproducibility-noinfo-details} gives the detailed \texttt{no\_info}-cohort transition counts. Across the ten \texttt{no\_info}-only reruns, only 1.4\% of all rerun labels and 2.5\% of released-name rerun labels switch to \texttt{found}. Most non-\texttt{no\_info} rerun labels are ambiguity signals (57.7\% of all rerun labels; 60.8\% for released names).

\input{filtered_recombination_country_pair_full_heatmap_figure.tex}
\input{release_candidate_country_pair_full_heatmap_figure.tex}
\input{accepted_unknown_country_pair_full_heatmap_figure.tex}

\subsection{Comparison-Name Construction Details}
\label{sec:appendix-comparison-names}

Table~\ref{tab:comparison-release-files} summarises the comparison-name release produced by the paper run. The combined \texttt{release/comparison\_names.csv} file contains the accepted unknown names, known-public-person controls, famous-name perturbations, and stable-\texttt{no\_info} subset rows. The ambiguous/similar controls are written to a separate file because they are rejected validation cases rather than accepted unknowns or known-public-person controls.

\begin{table*}[t]
\centering
\small
\resizebox{\textwidth}{!}{%
\begin{tabular}{llrl}
\toprule
File / stratum & Role & Rows & Construction note \\
\midrule
\textcolor{black}{\texttt{release/comparison\_names.csv}}: \texttt{unknown} & accepted operationally unknown names & 300 & fixed-size release from controlled-search verdicts \\
\textcolor{black}{\texttt{release/comparison\_names.csv}}: \texttt{comparison\_1}--\texttt{comparison\_5} & public-person controls & 300 & \(60\) rows per diagonal fame/prevalence level \\
\textcolor{black}{\texttt{release/comparison\_names.csv}}: \texttt{famous\_name\_proximity\_d1}--\texttt{d12} & famous-name perturbations & 3,600 & \(300\) unique variants per proximity distance level \\
\textcolor{black}{\texttt{release/comparison\_names.csv}}: \texttt{stable\_no\_info\_3} & stricter unknown subset & 49 & duplicate rows for names always \texttt{no\_info} in the three-replicate rerun panel \\
\textcolor{black}{\texttt{release/comparison\_names.csv}}: \texttt{stable\_no\_info\_10} & stricter unknown subset & 27 & duplicate rows for names always \texttt{no\_info} in the ten-replicate rerun panel \\
\textcolor{black}{\texttt{release/ambiguous\_name\_controls.csv}} & ambiguity controls & 100 & sampled candidates rejected through similar-name or attractor evidence \\
\bottomrule
\end{tabular}
}
\caption{Paper-facing comparison and control files. The two \texttt{stable\_no\_info} strata duplicate accepted unknown-name rows for analysis; they are not additional unique names.}
\label{tab:comparison-release-files}
\end{table*}

Known-public-person controls are built from a Latin-script subset of enriched Wikidata humans. The paper run used a 2,000-record candidate pool with complete required Wikipedia metadata after merging/fetching \texttt{wikipedia\_pageviews} and \texttt{wikipedia\_article\_length}. Exact-name prevalence is measured with the quoted query ``\texttt{"Full Name"}'' and reused from stored checks when the same backend/query record already exists. Records are rank-binned into five fame bins and five exact-prevalence bins. In diagonal mode, a record with fame-bin position \(f\) and prevalence-bin position \(p\) is assigned to one of five ordered comparison levels by averaging the normalized bin positions and clipping to the range \(1\)--\(5\). The final known-person control file contains \(60\) rows per comparison level.

Let \(V\), \(L\), \(G\), and \(K\) denote page views, article length, Wikipedia language count, and Wikidata sitelink count. The fame score is
\[
\begin{aligned}
S={}&0.45\log(1+V)+0.25\log(1+L)\\
&+0.20\log(1+G)+0.10\log(1+K).
\end{aligned}
\]
It is only a sampling proxy. Exact-name prevalence is stored separately so analyses can distinguish public-person prominence from the web footprint of the literal name string.

Famous-name perturbations are generated from public-person source names with deterministic seeded edits. Levels \(d1\)--\(d2\) contain close orthographic, abbreviation, or double-edit variants; levels \(d3\)--\(d7\) use increasing suffix and recombination changes; levels \(d8\)--\(d12\) are fragmentary or near-jibberish recombinations. Each row stores \texttt{source\_name}, \texttt{source\_qid}, \texttt{proximity\_distance\_level}, a textual distance label, raw and normalized edit distance, generation strategy, donor-part provenance, and the exact quoted prevalence count for the generated variant string. The release was generated with exact perturbation counts enabled, yielding \(300\) rows at every distance level.

%
\subsection{LSTM Name-Likeness Details}
\label{sec:appendix-lstm-name-likeness}

The LSTM audit is a string-form diagnostic, not evidence of operational unknownness. It asks whether accepted \textsc{pun} names occupy a plausible full-name region under a character-level model trained on ParaNames. The score is bits per character (BPC), so lower values indicate higher predictability under the name model. We compare accepted unknown names with public-person controls, famous-name perturbation controls, and fixed non-name controls. The main plots in Figures~\ref{fig:lstm-boxplots} and~\ref{fig:lstm-scatter} show that accepted \textsc{pun} names remain closer to public-person names than to non-name controls, while preserving the low exact-name search footprint required by the protocol.

The two-dimensional interpretation matters. Public-person controls are highly name-like but intentionally have indexed evidence; random strings are evidence-sparse but not name-like. Accepted \textsc{pun} names are useful because they combine low BPC with zero or near-zero exact-name prevalence under the documented validation run. BPC therefore supports the form-plausibility claim, while web-facing validation remains the evidential basis for operational unknownness.

\subsection{Tokenizer Footprint Details}
\label{sec:appendix-tokenizer-footprint}

Tokenizer fragmentation is an auxiliary audit of familiar-looking subword structure. For each string, we compute token pieces per character under 14 tokenizers and use the median score across tokenizers. Higher scores mean that the string is split into more pieces per character. This audit targets a different signal from the LSTM: tokenizers encode model- and corpus-specific subword inventories, while BPC measures character-level predictability under a name-trained model.

\begin{figure}[h]
  \centering
  \includegraphics[width=1.0\columnwidth]{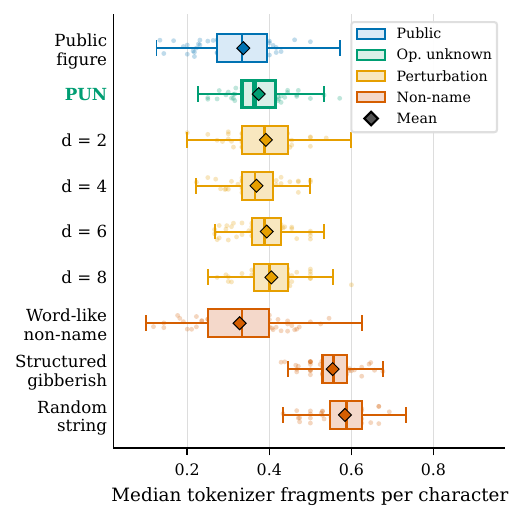}
  \caption{Median subword-tokenizer fragmentation across 14 tokenizers. Higher values indicate that tokenizers split the string into more pieces per character.}
  \label{fig:tokenizer-fragmentation-boxplots}
\end{figure}

\begin{figure}[h]
  \centering
  \includegraphics[width=1.0\columnwidth]{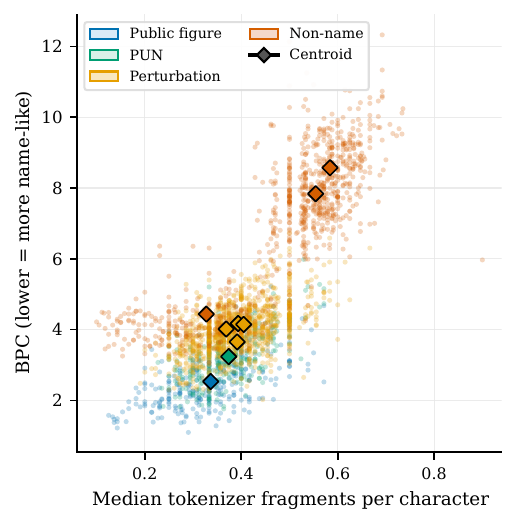}
  \caption{Tokenizer fragmentation compared with LSTM BPC.}
  \label{fig:tokenizer-fragmentation-scatter}
\end{figure}

Figures~\ref{fig:tokenizer-fragmentation-boxplots} and~\ref{fig:tokenizer-fragmentation-scatter} show that tokenizer fragmentation broadly agrees with the LSTM diagnostic. Accepted \textsc{pun} names have low median fragmentation, similar to public-person names and word-like non-names, and much lower than random strings. The correlation with BPC indicates shared sensitivity to string regularity, while the weak correlation with exact-name search prevalence confirms that tokenizer footprint is not a proxy for indexed full-name evidence.

\subsection{Ablation Details}
\label{sec:appendix-ablations}

The ablations were last run on 21 May 2026 from the state snapshot used for the paper release. Validation-stack ablations replay stored web-enabled LLM and controlled web search audit traces and therefore make no new web-facing validation calls. Table~\ref{tab:ablation-stage-attrition} reports stage attrition over the 52,726 generated candidates. Table~\ref{tab:ablation-serper-query} replays counterfactual controlled web search query policies for the 1,781 local-pass names that received a web-LLM \texttt{no\_info} label. Table~\ref{tab:ablation-generation-policy} reports 500-name diagnostic regeneration runs for generation-policy variants; these are not alternate releases, but checks on which generation constraints affect diversity and efficiency before validation.

\begin{table}[t]
\centering
\scriptsize
\begin{tabular}{lrrr}
\toprule
Stage & Category & Count & Pct. \\
\midrule
generated\_candidates & all & 52726 & 100.0 \\
local\_filter & pass & 49877 & 94.6 \\
local\_filter & rejected\_local & 2849 & 5.4 \\
azure\_web\_screen & similar & 27968 & 53.04 \\
azure\_web\_screen & fail & 9550 & 18.11 \\
azure\_web\_screen & attractor & 8129 & 15.42 \\
azure\_web\_screen & found & 2449 & 4.64 \\
azure\_web\_screen & no\_info & 1781 & 3.38 \\
serper\_verification & ambiguous & 1372 & 2.6 \\
serper\_verification & \shortstack[l]{no\_indexed\_full\_name\\evidence\_found} & 300 & 0.57 \\
serper\_verification & web\_evidence\_found & 108 & 0.2 \\
serper\_verification & validation\_incomplete & 1 & 0.0 \\
\bottomrule
\end{tabular}
\caption{Stage attrition in the validation stack. Percentages are relative to all generated candidates.}
\label{tab:ablation-stage-attrition}
\end{table}

\begin{table*}[t]
\centering
\scriptsize
\resizebox{\textwidth}{!}{
\begin{tabular}{lrrrrrrrr}
\toprule
Experiment & Evaluable Names & Accepted & Accepted Pct & Web Evidence Found & Ambiguous & Validation Incomplete & Newly Accepted Vs Full & Lost Full Stack Accepts \\
\midrule
full\_stack\_recorded\_verdict & 1781 & 300 & 16.84 & 108 & 1372 & 1 & 0 & 0 \\
exact\_quoted\_only & 1781 & 1115 & 62.61 & 92 & 573 & 1 & 815 & 0 \\
no\_unquoted\_queries & 1781 & 356 & 19.99 & 107 & 1317 & 1 & 56 & 0 \\
no\_fuzzy\_variants & 1781 & 942 & 52.89 & 93 & 612 & 134 & 642 & 0 \\
no\_page\_content & 1781 & 299 & 16.79 & 57 & 1396 & 29 & 0 & 1 \\
\bottomrule
\end{tabular}

}
\caption{Controlled-search query ablations on the web-LLM \texttt{no\_info} subset. \texttt{newly\_accepted\_vs\_full} counts names that would be accepted by the ablated query policy but are rejected by the full protocol.}
\label{tab:ablation-serper-query}
\end{table*}

\begin{table*}[t]
\centering
\scriptsize
\resizebox{\textwidth}{!}{
\begin{tabular}{lrrrrrr}
\toprule
Condition & Generated & Local Pass & Local Pass Pct & Same Country & Distinct Country Pairs & Attempts \\
\midrule
paper\_baseline & 500 & 468 & 93.6 & 0 & 243 & 22732 \\
allow\_same\_country & 500 & 472 & 94.4 & 31 & 241 & 20273 \\
pool\_size\_country\_sampling & 500 & 471 & 94.2 & 0 & 107 & 10052 \\
no\_rare\_token\_cap & 500 & 483 & 96.6 & 0 & 497 & 10590 \\
no\_native\_language\_requirement & 500 & 473 & 94.6 & 0 & 331 & 500 \\
allow\_source\_name\_match & 500 & 482 & 96.4 & 0 & 234 & 22398 \\
\bottomrule
\end{tabular}

}
\caption{Generation-policy ablations over 500 generated candidates per condition, followed by the same local validation profile.}
\label{tab:ablation-generation-policy}
\end{table*}

\subsection{Human Agreement Study Details}
\label{sec:appendix-human-agreement}

The human-audit study is run as a Prolific external study with a custom web application and database-backed packet assignment. Prolific passes participant, study, and session IDs through URL parameters. The application assigns the next packet with the fewest completed sessions and restores prior responses for returning participants. Stale started sessions are marked separately from completed sessions and are not counted toward packet completion.

Participants must consent, use a desktop/laptop browser, and pass a comprehension item about exact versus variant evidence. The Prolific screeners require English and access to at least one of the target chatbot providers. The study intentionally uses broad geographic distribution rather than country quotas; country-level demographics are obtained from Prolific metadata for reporting rather than for balancing.

Each packet contains six plausibility items, three fixed attention checks, six search tasks, one live chatbot task, and one saved-response classification task. The attention checks are drawn from three fixed control categories: a word-like non-name, structured gibberish, and a random character string. We exclude submissions with no consent, failed comprehension, duplicate participant IDs, completion under one third of the pilot median time, two or more failed attention checks, more than two unexplained ``could not check'' responses, or systematically unrelated pasted evidence. These exclusions are pre-specified and logged before final analysis.

\begin{figure}[t]
  \centering
  \includegraphics[width=0.9\columnwidth]{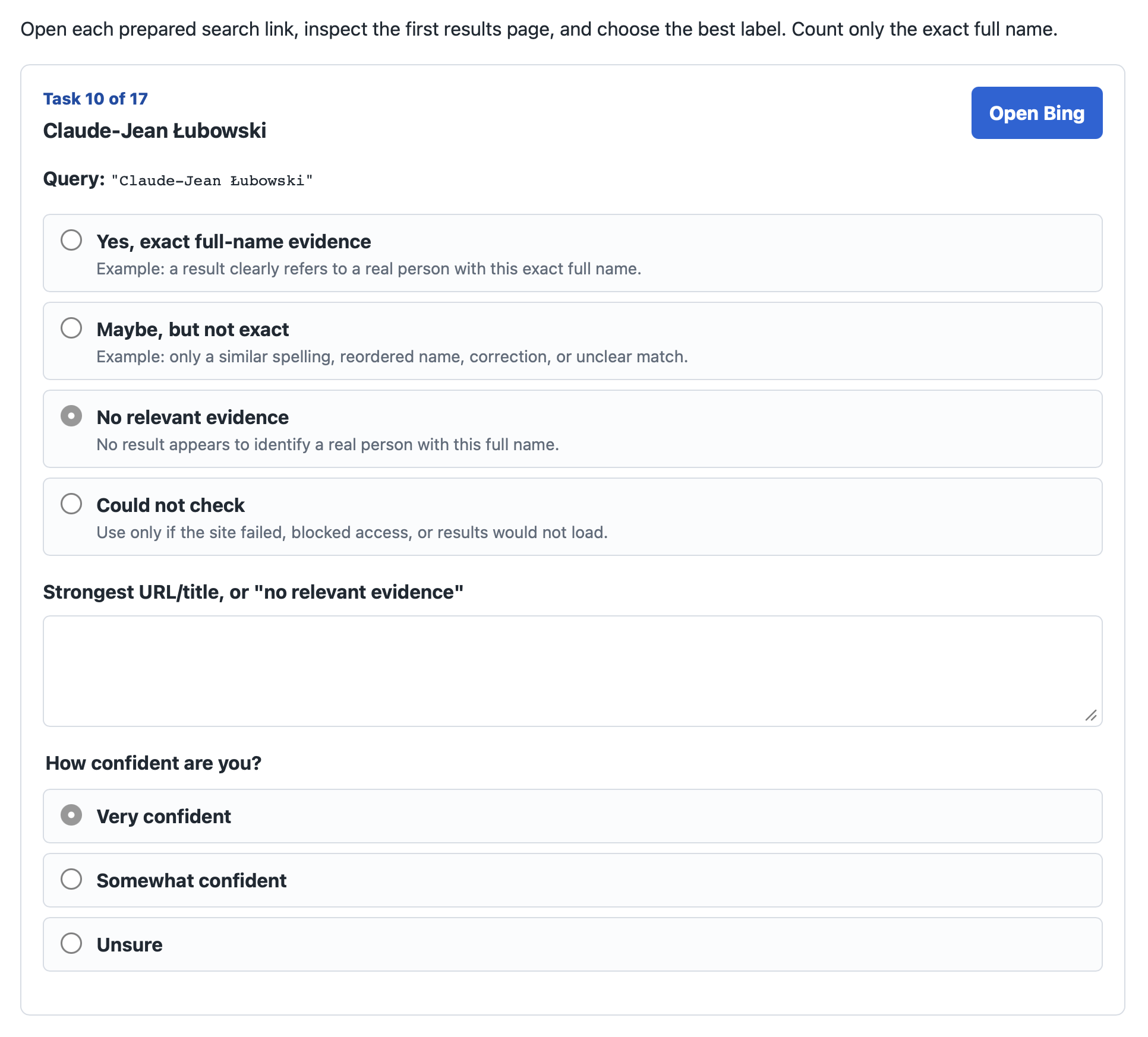}
  \caption{Controlled-search task screen with a prepared query link, exact-evidence labels, evidence field, and confidence item.}
  \label{fig:human-agreement-web-search}
\end{figure}

\begin{figure}[t]
  \centering
  \includegraphics[width=0.9\columnwidth]{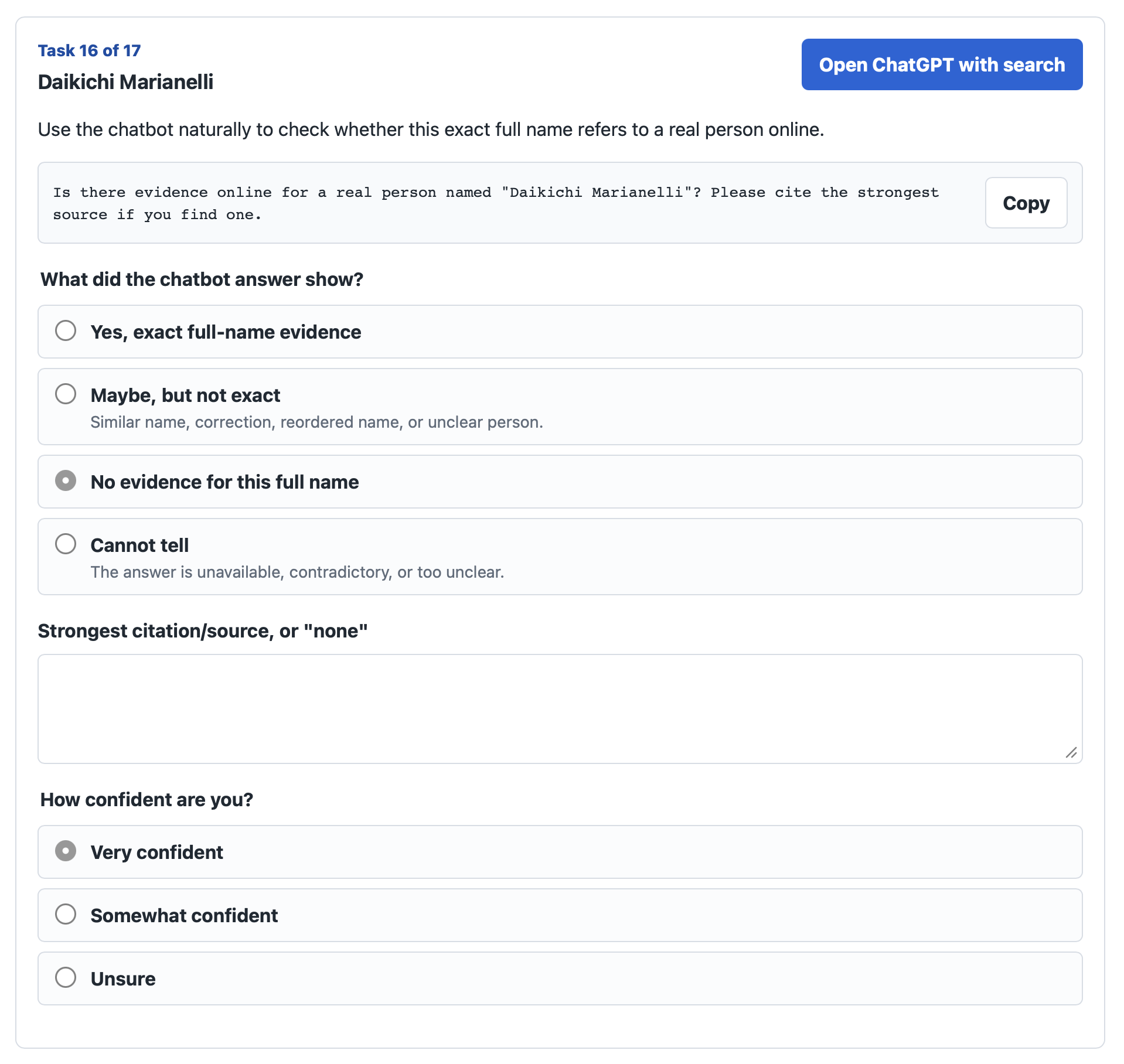}
  \caption{Live-chatbot task screen with a prepared prompt, exact-evidence labels, citation field, and confidence item.}
  \label{fig:human-agreement-chatbot}
\end{figure}

Figures~\ref{fig:human-agreement-web-search} and \ref{fig:human-agreement-chatbot} show the controlled-search and live-chatbot screens used in the human study.

The full export contains 224 completed submissions. After review exclusions, 205 completed submissions remain usable; for packet-level analysis we select the earliest non-flagged completed submission for each packet, yielding 204 selected participants and complete coverage of all 204 packets. Median completion time for the selected submissions is 8.2 minutes. The selected analysis set contains 1,224 plausibility judgments, 612 attention-check judgments, 1,224 controlled-search judgments, 204 live-chatbot judgments, and 204 saved-output classification judgments.

We inspect divergent evidence strings before interpreting raw disagreement as protocol error. Common failure modes include blank pasted evidence, text explicitly saying ``no relevant evidence'', search-engine result-page URLs, and snippets that contain only component-name or near-match evidence rather than the exact full-name string. The clearest informative divergences are ambiguity cases repeated across packets, e.g., \emph{\'Emile-Joseph Toshir\=o} triggering \emph{\'Emile Joseph} results, \emph{Karl-Peter Clarendon} triggering \emph{Carl Peters}, \emph{Kaitlyn Va\v{c}k\'a\v{r}} triggering \emph{Katie Vackar}, and \emph{Heinz-J\"urgen Frederick} triggering \emph{Heinz-J\"urgen Friedrich}. Reporting therefore separates raw participant labels, attention-filtered labels, and a conservative audited exact-evidence subset.

\subsection{Human-Subjects and Data Handling}
\label{sec:appendix-human-subjects}

The Prolific study was designed as a low-risk evidence-validation task. Participants did not make decisions about real people; they judged whether a displayed string looked like a full name and whether public search results or chatbot responses contained exact-person evidence, ambiguity signals, or no relevant evidence. The entry screen required consent, desktop/laptop access, and a comprehension check about exact-name versus variant evidence before assignment to a packet. Participants could report that a search could not be completed and could withdraw from participation at any time.

Participants were compensated through Prolific under the posted study terms. Payment review and analysis inclusion were kept separate: completed good-faith submissions were approved for payment even when excluded from analysis for pre-specified data-quality reasons. The exclusion full rules used for analysis were: no consent, failed comprehension, duplicate participant ID, completion under one third of the pilot median time, two or more failed attention checks, more than two unexplained ``could not check'' responses, or systematically unrelated pasted evidence.

\end{document}